\documentclass[runningheads]{llncs}

\usepackage{eccv}

\usepackage{eccvabbrv}

\usepackage{graphicx}
\usepackage{booktabs}

\usepackage[accsupp]{axessibility}  

\usepackage{xcolor}

\usepackage{hyperref}

\usepackage{orcidlink}

\usepackage{longtable}

\newcommand{\clip}[0]{\text{CLIP}}
\newcommand{\inputtext}[0]{\textbf{T}}
\newcommand{\bottleneck}[0]{\text{Diff-Bot}}
\newcommand{\unetoutput}[0]{\text{Diff-Out}}

\newcommand{\unet}{\text{U-Net}}
\newcommand{\pval}{P-value}

\newcommand{\modelsimilarity}[0]{\text{M-SIM}}
\newcommand{\datasimilarity}[0]{\text{H-SIM}}

\newcommand{\ttoi}[0]{\text{text-to-image }}
\newcommand{\Ttoi}[0]{\text{Text-to-image }}

\newcommand{\sota}[0]{\text{state-of-the-art }}

\begin{document}

\title{CLIP is All You Need for Human-like Semantic Representations in Stable Diffusion} 

\titlerunning{CLIP is All You Need}

\author{Cameron Braunstein\inst{1,2}
\and
Mariya Toneva\inst{2,3}
\and
Eddy Ilg\inst{3}
}

\authorrunning{C. Braunstein et al.}

\institute{Saarland University, Saarbrücken Germany \\
\email{braunstein@cs.uni-saarland.de} \and 
MPI for Software Systems, Saarbrücken Germany \and
University of Technology Nuremberg, Nuremberg Germany }

\maketitle

\begin{abstract}
  Latent diffusion models such as Stable Diffusion achieve state-of-the-art results on \ttoi{}generation tasks. However, the extent to which these models have a semantic understanding of the images they generate is not well understood. 
  In this work, we investigate whether the internal representations used by these models during \ttoi generation contain semantic information that is meaningful to humans. To do so, we perform probing on Stable Diffusion with simple regression layers that predict semantic attributes for objects and evaluate these predictions against human annotations. 
  Surprisingly, we find that this success can actually be attributed to the text encoding occurring in CLIP rather than the reverse diffusion process.
  We demonstrate that groups of specific semantic attributes
  have markedly different decoding accuracy than the average, and are thus represented to different degrees. 
  Finally, we show that attributes become more difficult to disambiguate from one another during the inverse diffusion process, further demonstrating the strongest semantic representation of object attributes in CLIP. 
  We conclude that the separately trained CLIP vision-language model is what determines the human-like semantic representation, and that the diffusion process instead takes the role of a visual decoder.
  \keywords{Diffusion \and Probing \and Generative Modelling \and Alignment}
\end{abstract}

\section{Introduction}
\label{sec:intro}

\Ttoi generation has undergone a recent, rapid advancement \cite{bie2023renaissance}. In particular, diffusion models produce \sota results on image generation conditioned on a text prompt~\cite{po2023state}. However, despite their success and the ability to steer the generation with text, the internal workings of diffusion models are not interpretable, and it is vastly unclear if the representations learned by such models align with human judgment. Work in the NLP domain has shown promising results that models aligned with human judgment can have improved performance, and are inherently more interpretable \cite{aw2023training,abdou2021does}, motivating us to study human alignment with vision models. In this work, we make a significant step towards better understanding \ttoi diffusion models by analyzing how their internal representations align with human perception.

For our analysis, we leverage well-known probing techniques~\cite{belinkov2021probing}. We use the MTurk dataset \cite{Sudre2012}, which consists of object nouns that serve as prompts, paired with ratings of over $200$ attributes associated to the objects. To probe whether the attributes are present similarly in the intermediate representations of the image generation process of Stable Diffusion~\cite{rombach2022highresolution}, we train linear ridge regressions to predict the attribute ratings. Since they constitute a simple mapping, their ability to accurately predict attributes tells us the extent to which the intermediate representations of Stable Diffusion align with human judgment. 

Our analysis reveals that the representations present in Stable Diffusion have the strongest alignment at the final layers of the CLIP~\cite{radford2021learning} text encoder. This comes as a surprise, as it indicates that semantic understanding comes mostly from the pretrained CLIP model and not from the reverse diffusion process. Instead, the reverse diffusion process can be seen as a visual decoding of the representation provided by CLIP. 

We provide a detailed analysis of how well certain groups of semantic attributes align with human annotation and which groups of attributes are represented well. Finally, we investigate how well Stable Diffusion can disentangle such attributes. To summarize, the contributions of this work are as follows:
\begin{itemize}
    \item We apply probing techniques to a \ttoi diffusion model, and show that these techniques are effective in the task of object attribute prediction for Stable Diffusion as an example of a generative model.
    \item We demonstrate that the semantic understanding in Stable Diffusion comes from CLIP instead of the diffusion model, and show that the reverse diffusion process acts as a visual decoding. 
    \item We provide a detailed analysis and show that object attributes extracted from CLIP align very well with human judgment, and furthermore, that CLIP is able to disentangle attributes which humans tend to closely associate. 
\end{itemize}

\section{Related Work}
\label{sec:prior_work}

\subsection{Text-to-Image Generative Models}

\Ttoi generative models have undergone rapid advancement in recent history, and for a comprehensive overview we refer to~\cite{zhang2023texttoimage}. Notable models, including DALLE \cite{ramesh2021zeroshot},  DALLE-2 \cite{ramesh2022hierarchical}, eDiff-I \cite{balaji2023ediffi}, Imagen \cite{saharia2022photorealistic}, and GigaGAN \cite{kang2023scaling}, produce impressive results, but are closed source and do not allow an analysis of the models. Open source alternatives are plentiful, including DiT \cite{peebles2023scalable}, GLIDE \cite{nichol2022glide}, and several generations of Kadinsky models \cite{arkhipkin2023kandinsky}. 

From these open source options, we chose Stable Diffusion~\cite{rombach2022highresolution} as a suitable reference work, and conduct our evaluations exclusively on this architecture. Stable Diffusion was trained on general-purpose data and a wide domain of images. It distinguishes itself from similar models by already having a rich literature on investigations on its interpretability (see \cref{subsec:prior_explainability}), which we seek to extend in our own work. Diffusion models work by drawing a data sample from a simple distribution (typically Gaussian noise), and denoising it into a sample from a more complex distribution, in this case, the distribution of plausible images \cite{luo2022understanding,po2023state}. Unlike earlier diffusion works \cite{sohldickstein2015deep}, which sample images in the RGB space, Stable Diffusion is the seminal work for performing the reverse diffusion process in a more efficient latent space that is obtained from a pre-trained VAE.
To produce images conditioned on text, a method for bridging the language and image modalities is required. Many state-of-the-art models \cite{rombach2022highresolution,kang2023scaling,ramesh2022hierarchical,balaji2023ediffi,nichol2022glide,li2023gligen} including Stable Diffusion achieve this using Contrastive Language-Image Pre-training (CLIP)~\cite{radford2021learning}, which consists of a language and a vision encoder that are trained to produce similar encodings for a caption and corresponding image on large amounts of unlabeled data.

\subsection{Prior Investigations of Explaining Stable Diffusion and CLIP}
\label{subsec:prior_explainability}
Prior works have investigated the interpretability of Stable Diffusion and of CLIP in isolation. For the most part, works that study the interpretability of Stable Diffusion or similar models do so with image editing or more accurate text conditioning \cite{zhang2023adding,brooks2023instructpix2pix,hong2023improving,kim2023dense,hertz2022prompttoprompt,li2023gligen} as the primary goal. One of these works, one that is closely related to ours is by Park \etal \cite{park2023understanding}, which explored manipulating Stable Diffusion's image latents for image editing. Their work is inspired by Kwon \etal \cite{kwon2023diffusion}, in which they demonstrate that diffusion models have a semantically interpretable latent space, and can adjust the latent in space to produce an intentional change in the semantics of the output image. In contrast, our work investigates latent interpretability by comparing it directly to human rated attributes. Our work is thus a novel investigation into the explainability of Stable Diffusion. 

Existing works have investigated CLIP's compositionality in multi-word prompts  \cite{lewis2023does,thrush2022winoground,ray2023cola,yun2023visionlanguage}, and CLIP's ability to rate image aesthetics \cite{Hentschel2022}. Schiappa \etal investigate CLIP's relational, attribute, and contextual understanding \cite{schiappa2023probing}. But, in their investigation, they look at whether CLIP can ununderstand attributes if they are passed as adjectives in the text prompt. Our investigation is novel for several reasons: unlike previous works, we investigate the CLIP text encoder's alignment with human perception of semantics on a single object. This is a more challenging task, as we are not querying CLIP with the attribute directly with the text prompt. An additional novelty is that we put CLIP's understanding of attributes into greater context by searching for alignment with directly with human perception.

\subsection{Interpretability with Probing}

Few works focus purely on interpreting Stable Diffusion's generative process, notably, Tang \etal \cite{tang2022daam} analyze where different words of a text prompt are expressed in an image. Our focus is on the interpretation of intermediate latent representations, to see if the model has a human-like understanding of the objects it generates. We use \textit{probing} as a tool to interpret the model.

Probing is a simple, effective technique for measuring AI model alignment with human perception \cite{muttenthaler2023human}. It emerged in the NLP domain with works by Köhn \cite{kohn2015}, Gupta \etal \cite{gupta2015}, and Shi \etal \cite{shi2016}, but relevant works in the computer vision field are by Alain and Bengio \cite{alain2018understanding}, and Muttenhaler \etal \cite{muttenthaler2023human}, which explored network alignment with humans on tasks such as odd-one-out classification and image classification. Our work applies these interpretability techniques to the novel domain of latent diffusion models, and explores alignment on the sophisticated task of object attribute understanding, which has not been done previously in computer vision. 
We elaborate on the technical details of probing in \cref{subsec:measuring_alignment}. Our work is a unique contribution to the interpretability of Stable Diffusion and especially CLIP. 



\section{Method}
\label{sec:method}

In \cref{subsec:background}, we create notation to precisely label the intermediate representations created by Stable Diffusion, which we use in our analyses. \cref{subsec:measuring_alignment} explains how model-human alignment is quantified with the technique of \textit{probing}, by first introducing general mathematical notation, and then plugging in our notation from \cref{subsec:background}. Finally, the concept of \textit{entanglement}, a measure for quantifying whether the model disambiguates between related attributes, is introduced in \cref{subsec:measuring_entanglement}.

\subsection{Background Notation}\label{subsec:background}
We analyze intermediate representations from Stable Diffusion during text-to-image generation, which relies on CLIP for conditioning the generation on text prompts. The pipeline utilizes the CLIP ViT-L/14 architecture, which consists of a tokenizer, followed by $12$ hidden encoder blocks. We denote the output of the CLIP text encoder from text input $\inputtext$ as 
\begin{equation}
    \clip(\inputtext) \, ,
\end{equation}
with $\clip_l(\inputtext)$ specifying CLIP's output at hidden layer $l$. Latent feature map generation in Stable Diffusion is realized through a time-conditional U-Net \cite{ronneberger2015unet} architecture. The initial latent feature map is initialized as Gaussian noise $\epsilon$. The U-Net is then applied repeatedly $50$ times, where it receives the latent feature map generated by the previous iteration and features from $\clip(\inputtext)$ injected into it at various levels via cross attention as input. We examine internal representations of the U-Net at every iteration, both at the bottleneck and the output. Work from Kwon \etal \cite{kwon2023diffusion} suggests that the bottleneck has more easily interpretable semantics, however, we also note that features from $\clip(\inputtext)$ are inserted before the bottleneck, and thus it may be strongly influenced by CLIP.  We write the bottleneck output at iteration $k$ as:
\begin{equation}
    \bottleneck_k( \clip(\inputtext)) \, ,
\end{equation}
and the U-Net output at iteration $k$ as:
\begin{equation}
    \unetoutput_k(\clip(\inputtext)) \, ,
\end{equation}
where we denote the final generated latent feature map with $\unetoutput(\clip(\inputtext))$. The generation is overall conditioned on $\inputtext$ and random noise, and the functions $\bottleneck_k$ and $\unetoutput_k$ produce  different outputs if the noise is changed. We use this to create multiple samples for a given input $\inputtext$. The latent feature map is passed into a decoder $\mathrm{Dec}$ to get the final output image $\mathrm{Dec}(\unetoutput(\clip(\inputtext)))$. An illustration of this architecture can be found in \cref{fig:method_overview}.

\subsection{Measuring Alignment}
\label{subsec:measuring_alignment}
We probe Stable Diffusion to understand the alignment between internal representations of a \ttoi generative model and human perception. In the following, we introduce our probing mechanism and then explain how it can be used to measure alignment. The input to the probing is a set of stimuli $\{s_i | i \in 1,...,n\}$, which in our case are text prompts that correspond to objects. We then pass the stimuli $s_i$ to the network and extract the outputs from different inner layers of CLIP, as well as at the bottleneck and output of the U-Net after each iteration of the reverse diffusion process. In general, we want to attach different probes to all of these intermediate representations to understand where certain attributes are present. However, in the following, we will first concentrate the discussion on a single probe of a network $f$ and define the output for stimuli $s_i$ as $x_i$:
\begin{equation}\label{eq:stimuli_def}
    x_i= f(s_i) \, .
\end{equation}
$s_i$ is also presented to humans to capture their rating $y_{i,j} \in \mathbb{R}$ across $j \in 1,...,m$ response classes. Probing argues \cite{belinkov2021probing} that the neural network and human are aligned for response class $j$, if there is a simple model (called a \textit{probe}), that can effectively predict $y_{i,j}$ from $x_i$, because a successful probe implies that the necessary information to predict $y_{i,j}$ is readily available in the neural network's representation. 

In our case, responses $y_{i,j}$ are scalar values that represent human annotator ratings and we can use a linear model as a probe. We denote the model weights as $\beta_j$ and $c_j$, and calculate the predicted scalar value:
\begin{equation}
    \hat{y}_{i,j} = x_i^T \beta_j + c_j \, .
\end{equation}
We write the vector of responses across all text prompts for a single attribute $j$ as $Y_j$, and the predictions across all text prompts as $\hat{Y}_j$:
\begin{equation}
    Y_j = (y_{1,j},y_{2,j}, \dots, y_{n,j} )\,\,, \hat{Y}_j = (\hat{y}_{1,j},\hat{y}_{2,j}, \dots, \hat{y}_{n,j} )  \, .
\end{equation}
To assess the performance of this model in the same units as the scalar $y_{i,j}$, we compute the root mean squared error (RMSE) across all stimuli: 
\begin{equation}
    \text{RMSE}(Y_j, \hat{Y}_j) = \sqrt{\frac{(Y_j - \hat{Y}_j )^T(Y_j - \hat{Y}_j )}{n}} \,.
\end{equation}
Leveraging ridge regression to estimate $\beta_j$ and $c_j$ can make the linear model more robust to unseen $x_i$ \cite[pp.60]{hastie01statisticallearning}, and we therefore formulate the following ridge regression problem:
\begin{equation}\label{eq:ridge_regression_original}
    \beta_j, c_j = \text{argmin}_{ \beta_j, c_j} \{ \alpha_j \cdot  \beta_j^T \beta_j + n \cdot \text{RMSE}((Y_j, \hat{Y}_j))^2 \} \, ,  
\end{equation}
where $\alpha_j \geq 0$ is a scaling hyperparameter on a regularization term, used to keep entries of $\beta_j$ from becoming too large and overfitting. Following existing conventions to improve the robustness of the regression \cite{Jeffers1967}, \cite[p.63]{hastie01statisticallearning}, we reduce the dimensionality of network outputs $x_i$ via PCA \cite{Jolliffe2016}, and then normalize their z-scores (\ie, we set the mean of each feature channel of $\{x_i\}$ to $0$, and the standard deviation of each feature channel of $\{x_i\}$ to $1$ \cite{han2012mining}).


\begin{figure}
  \centering
  \includegraphics[height=5.8cm]{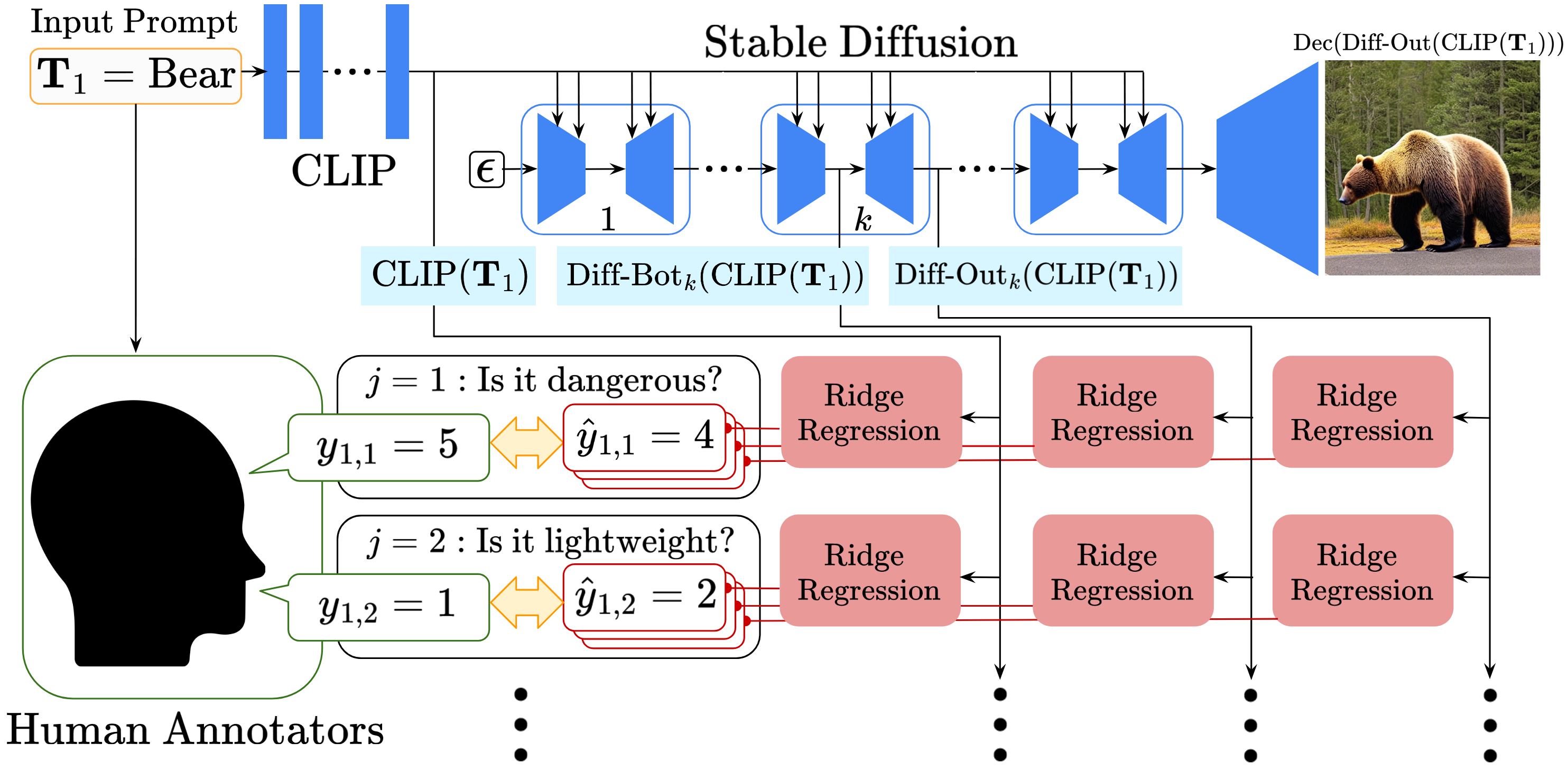}
  \caption{
  An overview of our probing method, focused on iteration $k$ of the latent generation. The stimulus text prompt $\inputtext_1$ (in this case, ``Bear'') is passed to Stable Diffusion. The intermediate object representation $\clip(\inputtext_1)$, $\bottleneck_k(\clip(\inputtext_1))$, and $\unetoutput_k(\clip(\inputtext_1))$ are being extracted from the model during the generation process of the image of a bear. For every attribute $j$, we would like to decode, each intermediate representation is passed to a unique ridge regression model that is trained to predict this attribute value. These predictions $\hat{y}_{i,j}$ are compared against the human annotator responses $y_{i,j}$, which are judgments about the attribute intensities for the object "Bear". Note that each ridge regression produces a unique set of predictions $\{\hat{y}_{i,j}\}$. In the diagram, the predictions for ridge regressions on $\clip(\inputtext_1)$ are shown on the front-most red boxes. Not all extracted intermediate representations have been shown: each $\clip_l (\inputtext_i)$, $\bottleneck_k(\clip(\inputtext_i))$, and $\unetoutput_k(\clip(\inputtext_i))$ for all $l$ and $k$ are extracted, and have their own ridge regressions. The model is tested on stimuli $\inputtext_i$ that have been withheld during training.
  }
  \label{fig:method_overview}
\end{figure}

Note that error measures by itself do not prove statistical significance and ridge regression is also not the only possible regression strategy~\cite{hastie01statisticallearning}. To address both of these concerns, we perform a permutation test on each ridge regression and report the \pval s \cite{Good2000}, which measure the likelihood that the regression's performance is only due to chance.
This also decouples our alignment result from the particular choice of ridge regression. 
Following Ojala \etal's approach~\cite{5360332}, we conduct our permutation test by computing the RMSE, but with permuted human responses:
\begin{equation} \label{eq:permuted_vector}
    \pi_p (Y_j) =  (y_{\pi_p(1),j},y_{\pi_p(2),j}, \dots, y_{\pi_p(n),j} ) \,
\end{equation}
and calculate the \pval{} from a collection of permutations $\Pi = \{\pi_p\}$ as:
\begin{equation}
    p = \frac{|\{ \pi_p \in \Pi \,|\, \text{RMSE}(\pi_p(Y_j), \hat{Y}_j) < \text{RMSE}(Y_j,\hat{Y}_j)\}| + 1}{|\Pi| +1 } \, .
\end{equation}
A sufficiently low $p$ then implies that the probe's success is statistically significant, \ie, the network has a meaningful understanding of attribute $j$.

For our probing of Stable Diffusion, our stimuli $\{s_i\}$ consist of single-word text prompts of concrete nouns $\{\inputtext_i\}$, all of which are common objects. The response $y_{i,j}$ is a human rating of the noun $\inputtext_i$ for an attribute $j$, and is an integer value between $1$ and $5$. 
For example, if $\inputtext_i$ is ``Bear'', and attribute $j$ is ``is it dangerous?'', then the human rating $y_{i,j}$ could be $5$, as bears are generally considered dangerous. Conversely, if attribute $j'$ is ``is it lightweight?'', $y_{i,j'}$ could be $1$, as bears are heavy.
As indicated in Figure~\ref{fig:method_overview}, to apply probing to Stable Diffusion, we now use separate probes  for each $\clip_\ell$, $\bottleneck_k$, and $\unetoutput_k$.


\subsection{Measuring Entanglement}\label{subsec:measuring_entanglement}
In addition to measuring the model-human perception alignment on each attribute individually, we also want to measure whether the complete collection of attributes are related to each other in the same way in both the model and human perception domains. 
We call this relationship \textit{entanglement}, and consider two attributes to be entangled in a domain if their representations are significantly similar, and disentangled if they are significantly dissimilar. It is now interesting to investigate the difference in entanglement in the model and human perception domain. 

The regression weights $\beta_j$ and $\beta_{j'}$ act as the representation for attributes $j$ and $j'$ in the model domain, as they will be high if certain features correlate with certain attributes consistently. The human annotation responses $Y_j$ and $Y_{j'}$ act as the attribute representation in the human perception domain. 
In both domains, we measure the attribute pair similarity by computing the cosine similarity for the model weights and for the human responses. After normalizing the channel-wise z-score of all regression weights $\{\beta_j \}$, or all attribute representations $\{Y_j\}$, we can compute the model and human perception similarities as follows:
\begin{equation}\label{eq:model_similarity}
    \modelsimilarity(\beta_j,\beta_j') = \frac{\beta_j^T \beta_{j'}}{||\beta_j || \cdot ||\beta_{j'}||} \, ,  \datasimilarity(Y_j,Y_{j'}) = \frac{Y_j^T Y_{j'}}{||Y_j || \cdot ||Y_{j'}||} \, .
\end{equation}

As in \cref{subsec:measuring_alignment}, we can quantify the significance of an entanglement via a permutation test, following the methodology presented by Ojala \etal \cite{5360332}. We will show how to carry this test out for the human perception similarity, although the implementation is isomorphic in the model case. We reuse the permutation notation defined in \cref{eq:permuted_vector} to calculate the \pval{} as:   
\begin{equation}
    p = \frac{|\{ \pi_n \in \Pi |  (\datasimilarity(\pi_n(Y_j),Y_{j'})) <  \datasimilarity(Y_j,Y_{j'})\}|+1}{|\Pi|+1 } \, .
\end{equation}
With \pval s we can quantify entanglement. If, in a given domain, the attribute pair's \pval{} is sufficiently high, we say the attribute pair is \textit{positively} entangled, and we expect the attributes to be semantically similar. If the \pval{} is sufficiently low, we say that the attributes are \textit{negatively} entangled, and we expect the attributes to be semantically opposite. Otherwise, we say that the attributes are \textit{disentangled}, and expect them to be semantically unrelated. We conclude our numerical analyses by looking at the changes in entanglement between domains in \cref{subsec:weight_entanglements}.

\section{Experiments}
\label{sec:results}

\subsection{Datasets}
\label{subsec:datasets}
To train and test our probes, we use a dataset of human annotations collected from the Mechanical Turk crowd-sourcing platform, which we refer to as the MTurk dataset~\cite{Sudre2012}. It consists of $1{,}000$ concrete nouns, $229$ attributes, and ground-truth ratings ranging from $1$ to $5$ for every object/attribute pair. Ratings in MTurk are the median rating from at least three human annotators. In contrast to~\cite{wang2022fmri}, this dataset provides significantly more nouns and attributes and in contrast to~\cite{mcrae2005semantic}, it provides direct integer ratings. 

Due to the large size of the training data for Stable Diffusion, which was trained on the LAION 400-M dataset with $400$ million image/text pairs~\cite{schuhmann2021laion400m}, and for CLIP, which was trained on the WebImageText dataset that also contains the same amount of image/text pairs~\cite{radford2021learning}, we do not expect a distribution shift from the MTurk data.

\subsection{Implementation Details}\label{subsec:implementation_details}
When we run Stable Diffusion, we use DDIM sampling \cite{song2022denoising} rather than DDPM sampling, as it is more computationally efficient. We sample Stable Diffusion $50$ times for each text prompt $\inputtext_i$, and treat each collected latent feature map $\bottleneck_k(\clip(\inputtext_i))$ and $\unetoutput_k(\clip(\inputtext_i))$ as a unique intermediate representation for probing. For robust probe results, we implement nested cross validation \cite{Refaeilzadeh2009} and report results taken across all outer folds, see the supplement for details. We conduct a grid search on two hyperparameters: the regularization $\alpha_j$ and the number of principal components of our regression inputs $x_i$. See the supplement for details. The Ridge regression parameters are calculated using Cholesky decomposition~\cite{Haddad2009} for a precise closed form solution. Throughout this work, we use $|\Pi|=2500$ permutations for permutation tests. Following convention, we use $p < 0.05$ as the significance threshold for the RMSE \pval. Likewise, when discussing entanglement, we say that an attribute pair is positively entangled if $p > 0.95$, and a pair is negatively entangled if $p < 0.05$.

\subsection{Alignment Between Stable Diffusion and Humans}\label{subsec:alignment_results}

\begin{figure}[tb]
  \centering
  \includegraphics[height=3cm]{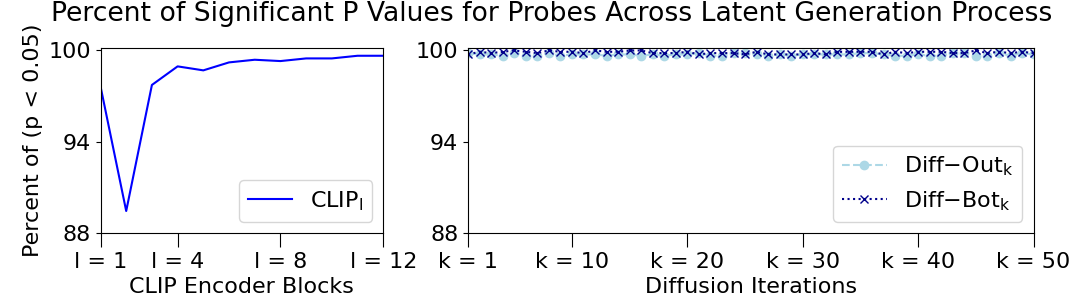}
  \caption{
  The percentage of significant predicted attributes with $p<0.05$ across all folds. 
  On the left, we visualize the percentages of $\clip_l$ probes. On the right, we visualize the percentages, both for probes of $\unetoutput_k$ and $\bottleneck_k$. We observe that most P-values are significant for probes across Stable Diffusion, with only a few non-significant ones across the hundreds of attributes that we probe. We provide a further analysis of the non-significant ones in the supplemental material.}
  \label{fig:p_values}
\end{figure}

\noindent\textbf{Visualizing P-Values.}
\label{subsec:p_values_full}
In~\cref{fig:p_values}, we visualize the P-values for alignment that we obtain from our probes.
We observe that $99.74 \%$ of the probes on the final output latent feature maps $\unetoutput(\clip(\inputtext_i))$ have a significant \pval. This is a striking result, as it shows that even the latent feature map that is decoded into an image has a general semantic representation of the objects it contains.
Furthermore, the performance of the ridge regression probes is overall significantly above chance, and we conclude that there is an alignment between Stable Diffusion's latent feature maps and the human perception of objects across a wide range of object attributes during the latent generation process.  

Even the output of CLIP exhibits a significant alignment for most attributes. As the CLIP features are inserted into each repetition and at various layers of the U-Net, it may also indicate that the alignment is actually induced by the CLIP encoder and not the diffusion model which is verified by the analysis in the following section. In the subsequent analyses, we only evaluate probes which achieved $p<0.05$, as we regard the outputs of the remaining probes as not meaningful.

\begin{figure}[tb]
  \centering
  \includegraphics[height=4.7cm]{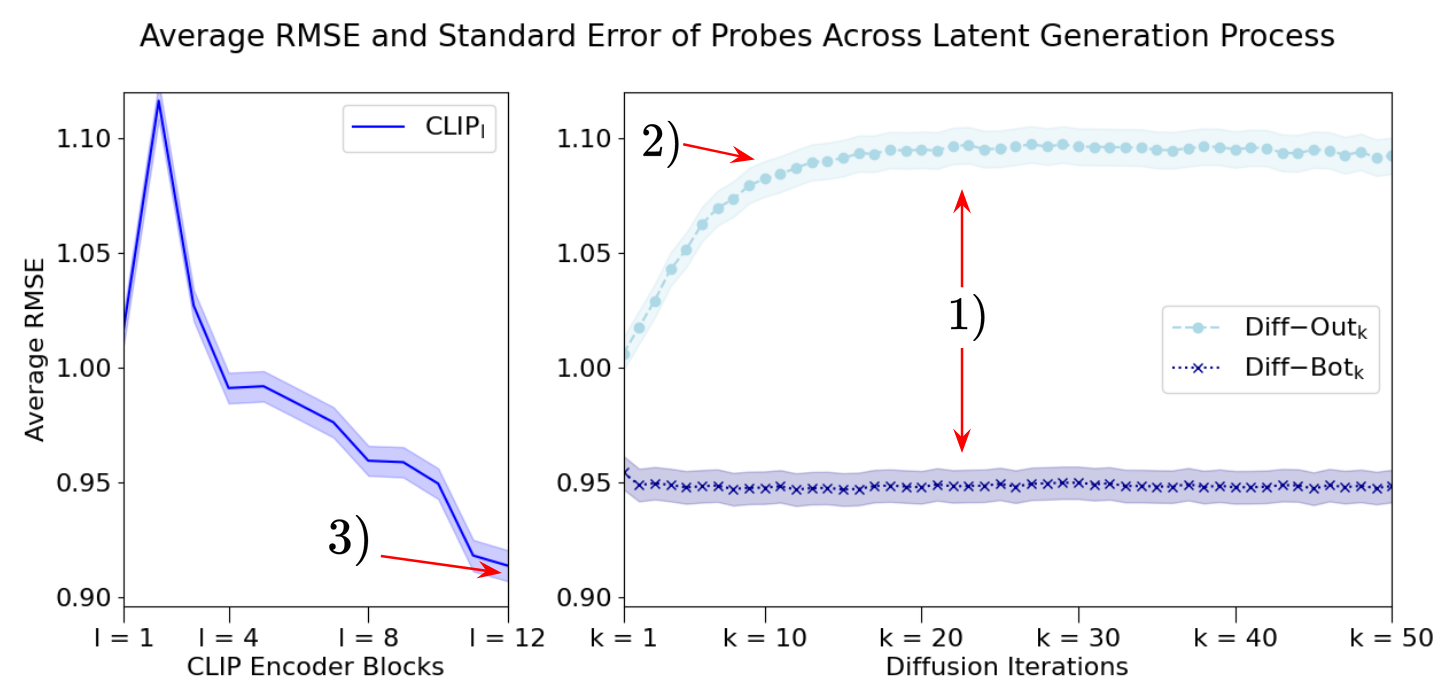}
  \caption{The average RMSE of the probes visualized with the standard error. Left, we show the RMSE of $\clip_l$ probes as a baseline. Right, we visualize the RMSE for probes of $\unetoutput_k$ and $\bottleneck_k$. Observations 1), 2), and 3) are elaborated on in the main text. }
  \label{fig:avg_rmse_over_network}
\end{figure}

\vspace*{2mm}
\noindent\textbf{Average RMSE and Standard Error.}
\label{subsec:avg_rmse} To visualize how close the predicted attribute ratings match the human annotations, we plot the RMSE across all attributes and outer folds over different stages of the model in \cref{fig:avg_rmse_over_network}. We observe the following: 1) The average RMSE is always lower for probes of $\bottleneck_k$ than for probes of $\unetoutput_k$ at every diffusion step $k$ and furthermore nearly constant. We conjecture that this is due to the CLIP features that are inserted at every iteration before the bottleneck. 
2) The average RMSE for $\unetoutput_k$ increases for probes at early iterations and then plateaus. This is expected, as the diffusion process converges from initial semantic concepts to pixel-level visual details. 
3) The average RMSE is lowest after the final layer of $\clip$. This finding indicates that the semantically most meaningful representation does not actually come from the diffusion model, but instead from the pretrained CLIP model. The diffusion model on the other hand only serves as a "visual decoding" of the representation provided by CLIP.
We verify that the RMSE minimum at $\clip$'s output is significantly lower than the RMSE across $\unetoutput_k$ and $\bottleneck_k$ using paired samples t-tests \cite{Xu2017-rq}. We find that the differences are significant ($p < 0.05$) across all $\unetoutput_k$ and $\bottleneck_k$. We conclude that the semantic representation of an object in Stable Diffusion is in fact most human-like at the output of $\clip$. Critically, the reverse diffusion process degrades this alignment between representations. 

\begin{figure}[t]
  \centering
  \includegraphics[height=7cm]{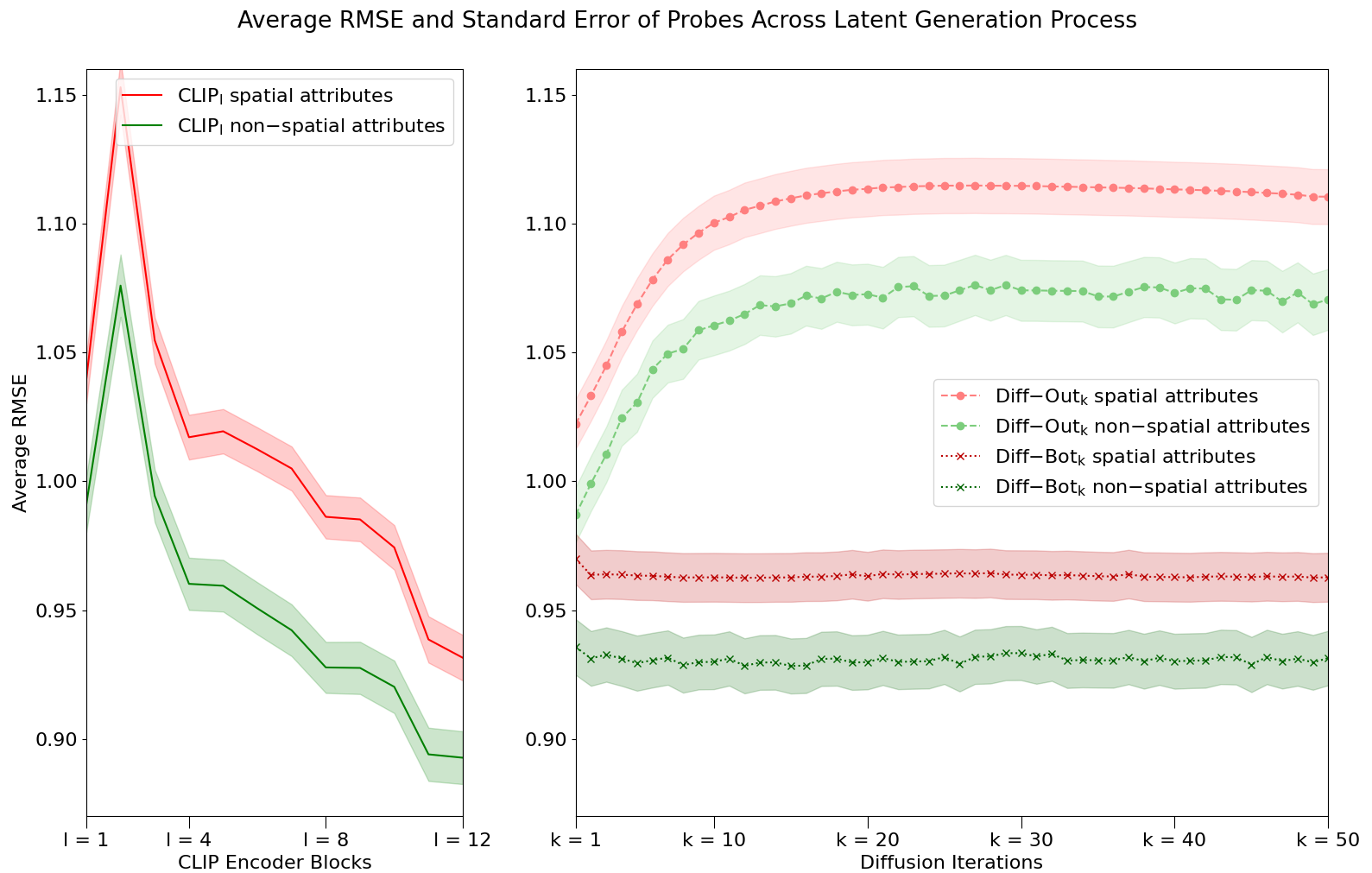}
  \caption{We visualize the average RMSE and standard error of all spatial (\textcolor{red}{red}) and non-spatial (\textcolor{OliveGreen}{green}) attributes. Spatial attributes have a higher average RMSE across all $\clip_l$, $\unetoutput_k$, and $\bottleneck_k$.}
  \label{fig:spatial_non_spatial}
\end{figure}

\noindent\textbf{Spatial and Non-Spatial Attribute Analysis.}
\label{subsec:spatial_non_spatial_analysis} Having demonstrated the effectiveness of our probing method in general, we want to explore whether certain groupings of attributes are more or less predictable. Of special interest to us is the difference between \textit{spatial} and \textit{non-spatial} attributes. Here, spatial attributes describe anything related to physicality or appearance of an object, for example: ``does it have corners?'', and non-spatial attributes are the remaining, for example: ``do you love it?''. For a full list of these attributes, see the supplement. We care about this distinction because we expect a model which generates images to have a better understanding of spatial attributes. Comparisons between spatial and non-spatial attributes are shown in \cref{fig:spatial_non_spatial}. On average, the non-spatial attributes are more decodable than the spatial attributes across all $\clip_l$, $\unetoutput_k$, and $\bottleneck_k$. This is unexpected, as Stable Diffusion is trained to produce an image latent feature map, so we expect spatial attributes to be more accurately decoded by probing. Additionally, in the supplement, we examine finer-grained subgroups for further insights.

\subsection{Weight Entanglements}
\label{subsec:weight_entanglements}

We analyze entanglement of $\clip_l$, $\unetoutput_k$, and $\bottleneck_k$, with the aim of understanding how entanglement changes between the model and human perception domains. To do this, we look at two quantities in particular: the first being the amount of attribute pairs which are entangled by the probes, but are disentangled by humans. The second being the amount of attribute pairs which are entangled by humans, but disentangled by the probes. See \cref{tab:p_value_changes} for our numerical results.

From these results, we conclude that $\clip$ can effectively disentangle attributes which humans entangle. The opposite case occurs across $\unetoutput_k$. Here, it is more common for attribute pairs which are disentangled by humans to become entangled in our models. We argue that the image latent feature maps do not effectively disambiguate related attributes. Given the relatively high RMSE for the $\unetoutput_k$, this suggests that attributes are generally less interpretable in the latent feature maps than in other regions of the model. Across $\bottleneck_k$, there are more attribute pairs which are entangled by humans that become disentangled in our models than attribute pairs which are disentangled by humans and entangled in our models, although the difference is not a pronounced as in CLIP. So, in the bottleneck of the \unet, attribute pairs are effectively disambiguated, although not as strongly as in CLIP. This finding supports our comments in \cref{subsec:avg_rmse}, that the U-Net bottleneck is more semantically interpretable than the output latent feature map. 

Generally, as an object representation passes from the text prompt to the image latent feature map, attributes become more entangled in the probe models. As the model generates an image from text, the object semantics may become less pertinent to the model as the representation shifts to the visual pixel space. 

\begin{table}
    \centering
    \begin{tabular}{|c|c|c|c|}
        \hline
          & Humans Disentangle & Probes Disentangle & Agreement between    \\
          & more than Probes & more than Humans & Humans and Probes \\
        \hline
        $\clip_l$ & \textcolor{red}{\textbf{3.7 \%}} &  \textcolor{blue}{\textbf{31.5\%}} & 64.8 \%    \\
        \hline
        $\bottleneck_k$ & \textcolor{red}{\textbf{10.1 \%}} &  \textcolor{blue}{\textbf{19.0 \%}} & 70.9 \% \\
        \hline
        $\unetoutput_k$ & \textcolor{red}{\textbf{20.2 \%}} &  \textcolor{blue}{\textbf{4.0 \%}} & 75.8 \% \\
        \hline
    \end{tabular}
    \vspace{2mm}
    \caption{We compare how attribute pairs are disentangled between the human ratings and probes across $\clip_l$, $\unetoutput_k$, and $\bottleneck_k$ for the outer folds of nested cross validation. We compare the total percentage of attribute pairs that are disentangled by humans and entangled in the probe weights (marked in \textcolor{red}{red}) with the total percentage of attribute pairs that are entangled by humans and disentangled in the probe weights (marked in \textcolor{blue}{blue}). In the final column, we list the percentage of attribute pairs that agree in both domains, that is, the attributes that are entangled by both, humans and the probes, or the attributes that are disentangled by both, humans and the probes. The final column is the remainder from the previous columns. For $\clip_l$, the \textcolor{blue}{blue} percentage is much higher than the \textcolor{red}{red} percentage. This difference is smaller for $\bottleneck_k$. For $\unetoutput_k$, the drastically more attribute pairs become entangled from the human to the model than vice versa. The overall observation from the last column is that all models agree with humans to a large degree in entangle- and disentanglement. 
    }
    \label{tab:p_value_changes}
\end{table}

    \vspace*{-5mm}

\subsection{Discussion and Future Direction}\label{subsec:discussion}
Our investigation shows that Stable Diffusion's reverse diffusion process does overall not improve semantic understanding much more than what can be obtained from CLIP. This is an interesting finding, as it indicates that the diffusion process does not learn semantics, but instead serves only as a visual decoder of the representation already available. Notably, diffusion models for image generation without language conditioning also exist and are capable of generating high-quality images. A future step in our investigation could be to apply our technique to those models next, to see if they exhibit any alignment with humans. However, to accomplish this requires a dataset of images and attribute labels that is not available today. We plan to create such a dataset in the future. This will also allow to not only compare attributes that are associated to general prompts, but to actually generated images. 

An additional finding from~\cref{subsec:avg_rmse} is that spatial attributes are less-well represented than non-spatial ones. This indicates that although Stable Diffusion is trained to output 2D images of the scenes, its training process does not learn good representations for spatial relationships and motivates future research on designing models that bring in such spatial relationship explicitly either in 2D or 3D. 

On the other hand, our results indicate that CLIP is able to disentangle attributes better than humans in many cases, which indicates that unsupervised training of large vision-language models is a promising approach to learn semantics that do not involve, or involve only limited spatial understanding. Future research should also investigate how well large language-only and vision-language models can align with human perception, respectively. 

As a next step, it will also be interesting to apply our technique to a variety of models in general, including more text-to-image diffusion models as well as GANs, to understand their respective differences in human alignment and reveal any favorable architecture biases.

\section{Conclusion}
\label{sec:conclusion}
In this work, we have explored the alignment between Stable Diffusion's latent representations of objects and human perceptions. We found that most human attribute ratings can be predicted from the model representations with an accuracy significantly below chance, and that CLIP is primarily responsible for generating these decodable representations. Not only are CLIP's object representations more decodable, but they are also more disentangled than those later in the generation process. In general, non-spatial attributes are in average more accurately decoded than spatial ones. The most salient insight from this analysis is that despite being a model trained to generate images of physical objects, conceptually high-level semantic attribute probes are more accurate than attributes related to physicality. In the future, we aim to create an image dataset labeled with attribute ratings and hope to generalize our results by probing a wide range of generative models. Our work represents a step towards documenting an alignment between generative models and human perception, that in the long term we hope will enable us to design AI models that are more in line with the human understanding of the world than today. 

\section{Acknowledgements}
Funded by the Deutsche Forschungsgemeinschaft (DFG, German Research Foundation) -- GRK 2853/1 “Neuroexplicit Models of Language, Vision, and Action” - project number 471607914. We would like to thank Tom Fischer and Raza Yunus for their feedback on this work. The authors gratefully acknowledge support from MPI for Software Systems.

\clearpage

\appendix
\section{Supplementary Material}

In \cref{subsec:nested_cv}, we provide details of the nested cross validation procedure promised in Sec. 4.2.
In \cref{subsec:grid_search}, we provide details of the grid search promised in Sec. 4.2.
In \cref{subsec:further_subgroup_analyses}, we provide the analysis of smaller subgroups promised in Fig. 2. 
In \cref{subsec:exceptional_attributes}, we provide an analysis of attributes whose probes had non-significant \pval{}s at the output of CLIP. In \cref{subsec:ethics}, we address the methods used to acquire the MTurk dataset. Finally, in \cref{subsec:attribute_subgroups}, we list the attributes used in subgroups, both in Sec 4.3,
and \cref{subsec:further_subgroup_analyses}.

\subsection{Nested Cross Validation Details}
\label{subsec:nested_cv}

Nested Cross Validation \cite{cawley2010,Stone1974} is a robust method for assessing models, which we use in our method. We divide our human annotation data into $5$ outer folds, each consisting of $200$ objects and all of their attribute annotations from the MTurk data. Each outer fold has ridge regressions trained on a regression regularization parameter $\alpha_j$, and a number of principal components, that have been optimized via cross validation for RMSE on the other $4$ outer folds. 

For this cross validation, the $4$ outer folds are treated as inner folds. We have regressions trained on each subset of $3$ inner folds, and validated on the remaining inner fold. This means that, for every $\alpha_j$ and principal component count, we have RMSE evaluated on $4$ models. We select the $\alpha_j$ and principal component configuration with the lowest average RMSE across these $4$ regression models through a grid search (see details in \cref{subsec:grid_search}), and which is used as the configuration for the ridge regression for the outer fold.

The ridge regression model is trained on the $4$ outer folds used during the cross validation, and is evaluated on the remaining outer fold. RMSE values which we report in our results section are the average across the evaluations of all the outer folds. Percentages of \pval{}s are also calculated across all outer folds.
\subsection{Grid Search Details}
\label{subsec:grid_search}

To find optimal hyperparameters for the inner fold cross validation, we conduct a grid search across the ridge regression regularizer $\alpha$, and the number of principal components, optimizing for the cumulative root mean squared error (RMSE) of the ridge regression models when assessed against the validation data.

Due to the high computation effort of a unique grid search across each regression model $\beta_j$, $c_j$, we search for hyperparameters which worked well across all $\clip_l$, all $\unetoutput_k$, and all $\bottleneck_k$, respectively. We argue that each of these components have a relatively similar representation space across $l$ or $k$, and therefore results will be close to the fine-grained grid searches. Furthermore, optimizing over the average RMSE for all attributes, rather than optimizing over each attribute individually makes our analysis of the alignment of Stable Diffusion more robust, as we are not over optimizing each attribute.  

For example, on the grid search for $\clip_l$, we compute the average RMSE for a hyperparameter configuration across all attributes, and all indices $l$.
In the case of $\unetoutput_k$ and $\bottleneck_k$, we evaluate only across every $10$th layer for computational efficiency.

\begin{figure}[tb]
  \centering
  \includegraphics[height=8.7cm]{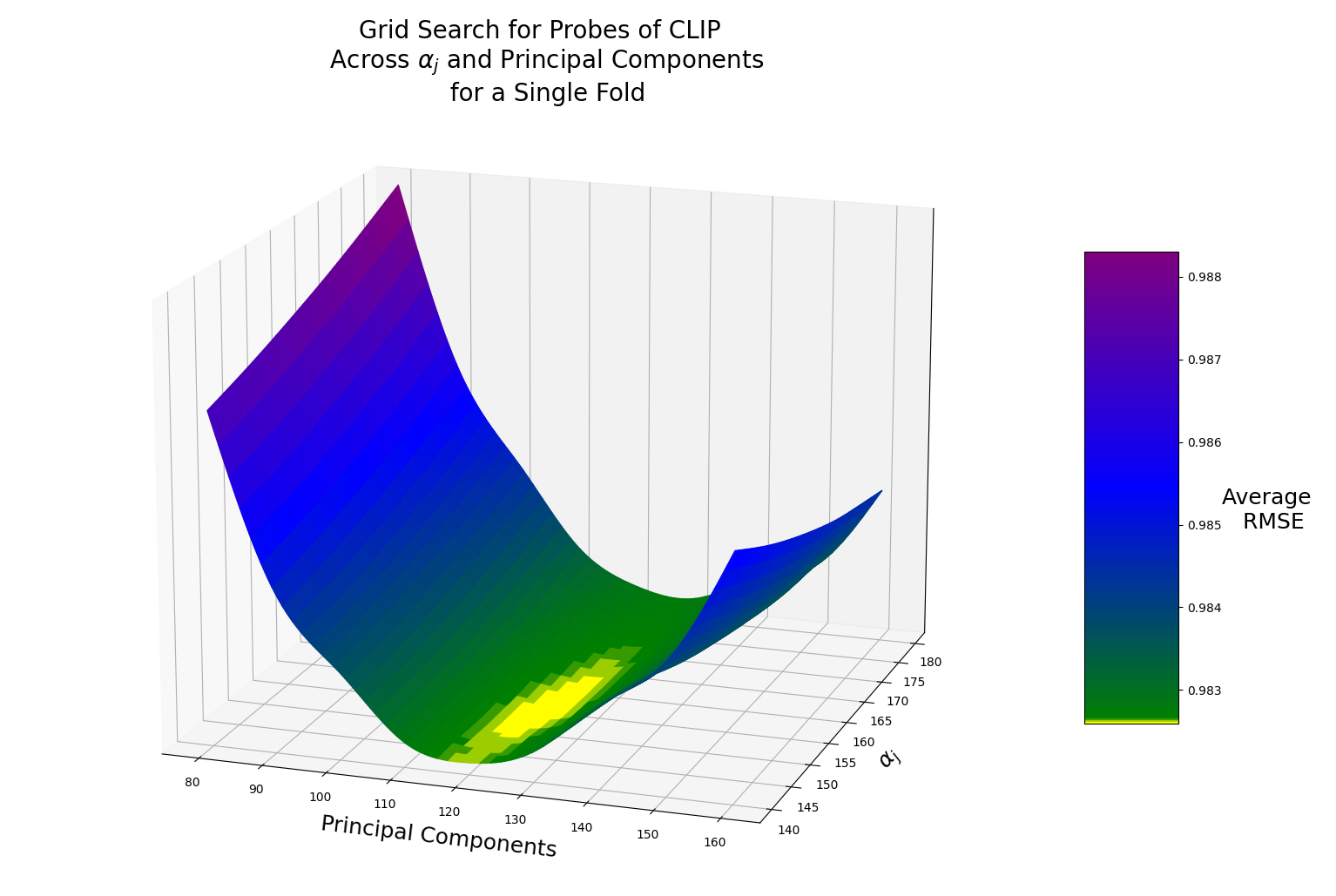}
  \caption{We visualize the grid search for the ridge regression hyperparameters for probes of CLIP for a single fold. The average RMSE has a saddle point near $120$ principal components, with $a_j = 150$. Therefore, we use these hyperparameters for the probes that are evaluated in our work. }
  \label{fig:grid_search_example}
\end{figure}

For $\clip_l$, our grid searches are run for $\alpha_j$ values between $110$ and $180$, and a number of principal components between $80$ and $160$. For $\unetoutput_k$, our grid searches are run for $\alpha_j$ values between $8{,}000$ and $14{,}000$, and a number of principal components between $1{,}550$ and $1{,}950$. For $\bottleneck_k$, our grid searches are run for $\alpha_j$ values between $5{,}000$ and $7{,}000$, and a number of principal components between $600$ and $1{,}100$.
We provide a visualization of this hyperparameter grid search for all $\clip_l$ in one fold of in \cref{fig:grid_search_example}

\subsection{Further Subgroup Analyses}\label{subsec:further_subgroup_analyses}

We extend our analysis in \cref{subsec:avg_rmse}, to understand probe performance over more focused subgroups of attributes \textemdash \ie attributes describing animacy, attributes describing perceptual features, and attributes describing size. Full lists of the attributes in these groups are in \cref{subsec:attribute_subgroups}. For each of these subgroups, we compare the average RMSE of our probes across the latent generation process for attributes within a subgroup, against all remaining attributes. We show and discuss the results for each of these subgroups below.

\begin{figure}[ht]
  \centering
  \includegraphics[height=6.7cm]{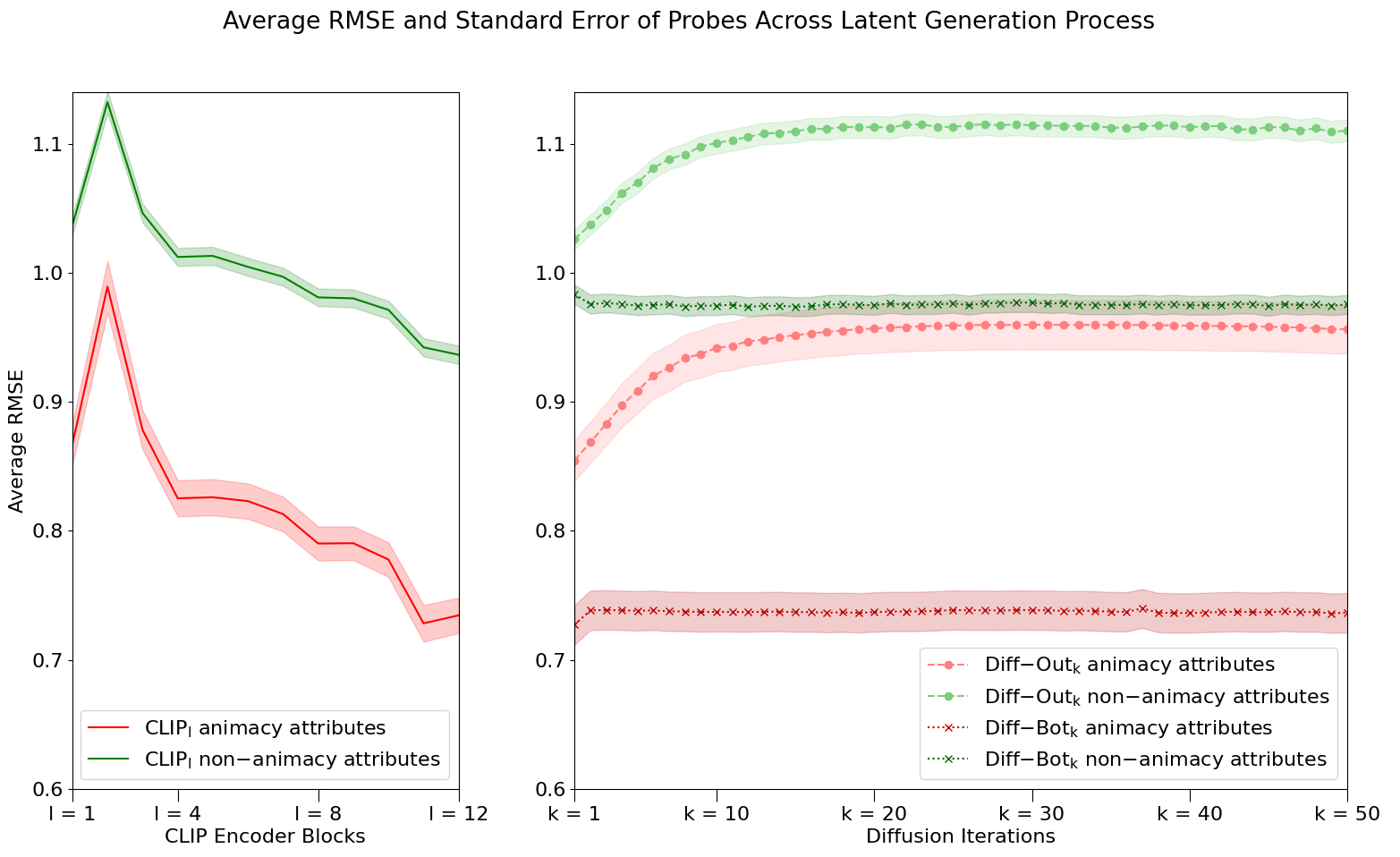}
  \caption{We visualize the average RMSE and standard error of all animacy attributes (\textcolor{red}{red}), and the remaining attributes (\textcolor{OliveGreen}{green}). Animacy attributes have a lower average RMSE across all $\clip_l$, $\unetoutput_k$, and $\bottleneck_k$.}
  \label{fig:animacy_rmse}
\end{figure}

\noindent{\textbf{Animacy.}} 
Animacy attributes are rated highly for living things. See \cref{fig:animacy_rmse} for the results. Attributes relating to the animacy of an object have lower RMSE across all components than the average. This suggests that Stable Diffusion maintains a strong understanding of animacy across the entire generation. The attributes describing perceptual features have higher RMSE than the average across all viewed components. 

\begin{figure}[ht]
  \centering
  \includegraphics[height=4.7cm]{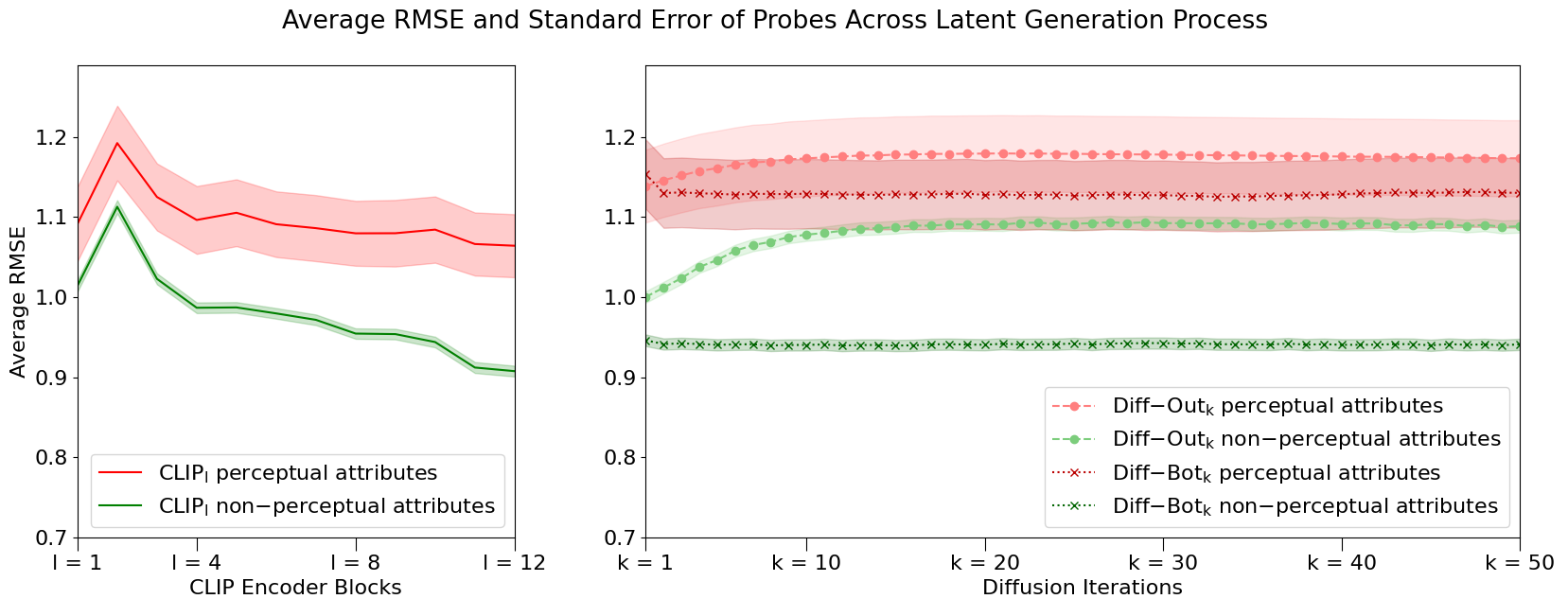}
  \caption{We visualize the average RMSE and standard error of all perceptual attributes (\textcolor{red}{red}), and the remaining attributes (\textcolor{OliveGreen}{green}). Perceptual attributes have a higher average RMSE across all $\clip_l$, $\unetoutput_k$, and $\bottleneck_k$.}
  \label{fig:perceptual_rmse}
\end{figure}

\noindent{\textbf{Perceptual.}}
Perceptual attributes describe low level visual features. See \cref{fig:perceptual_rmse} for the results. Perceptual attributes describe global spatial properties, which may not be semantically meaningful, and hence not present in the \unet{} bottleneck or in CLIP. Such a global spatial property may also not be easily interpreted by a linear predictor.

\begin{figure}[ht]
  \centering
  \includegraphics[height=5.7cm]{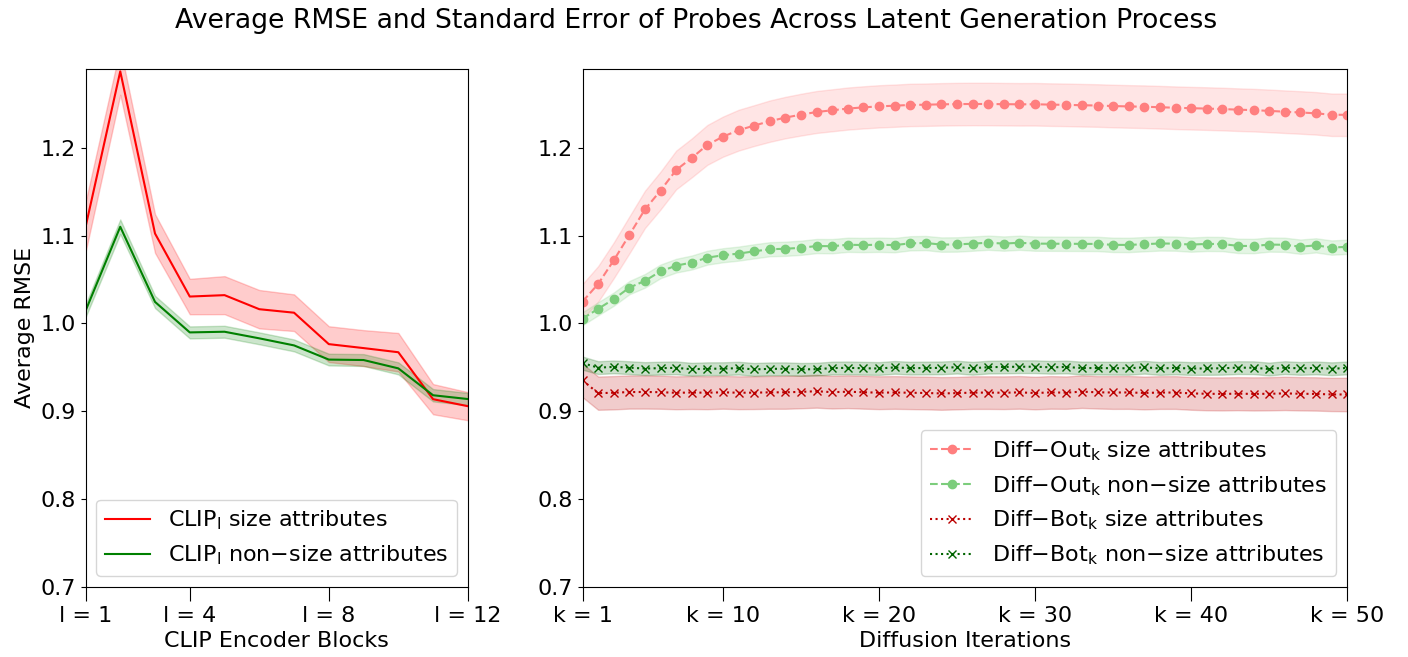}
  \caption{We visualize the average RMSE and standard error of all size attributes (\textcolor{red}{red}), and the remaining attributes (\textcolor{OliveGreen}{green}). Size attributes have a lower average RMSE across all $\bottleneck_k$, but a higher average across all $\unetoutput_k$ and most $\clip_l$.}
  \label{fig:size_rmse}
\end{figure}

\noindent{\textbf{Size.}}
 Size attributes describe the physical size of the object. See \cref{fig:size_rmse} for the results. We observe that size-related attributes have lower error than average for the later layers of the CLIP encoder and in the bottleneck of the \unet, but have higher than average RMSE for the \unet{} output. We conjecture that because size is a global object property rather than a local one, it is easily decodable in the compact representation of the \unet. However, it becomes challenging for the linear model to decode this property in the \unet{} output, as it is not expressed anywhere locally in the representation.

\subsection{Attributes with Non-Significant P-Values}\label{subsec:exceptional_attributes}

We examine the probes on the output of CLIP, $\clip(\inputtext)$ which fail the permutation test.
The attribute probes which we found to not have a significant \pval{} for at least one fold of evaluation were:  ``Is it a person?'', ``is it dense?'', ``is it a specific gender?'', and ``does it have feathers?''. 

The distribution of ratings for the attributes ``Is it a person?'', ``is it a specific gender?'', and ``does it have feathers?'' are extremely unimodal, that is, they have one value which occurs much more often that all others. This type of distribution can make regression models ineffective, as they are rewarded for setting $\beta_j$ nearly to $0$ and $c_j$ to the mode of the distribution. This can make the regression more susceptible to outliers, and prone to a high \pval{} in the permutation test.

The one attribute with a more even rating distribution whose regressions are still failing the permutation test is ``is it dense?''. We think this attribute is difficult to predict from CLIP because it depends on the ratio between the weight and size of an object. While the size of an object may be clear from an image, the weight may be difficult to understand without a deeper understanding of the world's physics. CLIP may have an approximate understanding of weight by equating it to size, but this could actually make its understanding of density (the ratio between weight and size), less accurate. Additionally, density may not be mentioned in image captions too frequently, meaning that neither modality that CLIP is trained on would have a strong notion of density. 

In $\bottleneck$ and $\unetoutput$, the only additional attribute which fails the permutation test at some inverse diffusion iterations is ``is it an insect?''. In this case, the distribution is also extremely unimodal, which can help explain why the permutation test failed.

\subsection{Crowd Sourcing Details}\label{subsec:ethics}
The MTurk dataset \cite{Sudre2012} was crowdsourced. It does not contain any information which may identify participants. For a wider conversation on the ethics of MTurk, we refer the reader to \cite{Moss2023}.

\subsection{Attribute Subgroups}
\label{subsec:attribute_subgroups}

In this subsection, we enumerate the attributes present in the MTurk dataset, and the subgroups which we created for our analyses in \cref{subsec:further_subgroup_analyses}. We informally looked through the attributes, and found subgroups pertaining to animacy (\cref{tab:animacy}), size (\cref{tab:size}), and perceptual features (\cref{tab:perceptual}).

\begin{center}
    \begin{longtable}{c}
        \caption{Animacy attributes} \label{tab:animacy} \\
        \hline \endfirsthead
            \hline
            Does it have a tail? \\ 
            \hline
            Does it have legs? \\ 
            \hline
            Does it have four legs? \\ 
            \hline
            Does it have feet? \\ 
            \hline
            Does it have paws? \\ 
            \hline
            Does it have feathers? \\ 
            \hline
            Does it have some sort of nose? \\ 
            \hline
            Does it have a hard nose/beak? \\ 
            \hline
            Can it run? \\ 
            \hline
            Is it fast? \\ 
            \hline
            Can it fly? \\ 
            \hline
            Can it jump? \\ 
            \hline
            Can it float? \\ 
            \hline
            Can it swim? \\ 
            \hline
            Can it dig? \\ 
            \hline
            Can it climb trees? \\ 
            \hline
            Can it cause you pain? \\ 
            \hline
            Can it bite or sting? \\ 
            \hline
            Does it stand on two legs? \\ 
            \hline
            Is it wild? \\ 
            \hline
            Is it a herbivore? \\ 
            \hline
            Is it a predator? \\ 
            \hline
            Is it warm blooded? \\ 
            \hline
            Is it conscious? \\ 
            \hline
            Does it have feelings? \\ 
            \hline
            Is it smart? \\ 
            \hline
    \end{longtable}
\end{center}

\begin{longtable}{c}
    \caption{Size attributes} \label{tab:size} \\
        \hline \endfirsthead
        \hline
        Is it smaller than a golfball? \\ 
        \hline
        Is it bigger than a loaf of bread? \\ 
        \hline
        Is it bigger than a microwave oven? \\ 
        \hline
        Is it bigger than a bed? \\ 
        \hline
        Is it bigger than a car? \\ 
        \hline
        Is it bigger than a house? \\ 
        \hline
        Is it taller than a person? \\ 
        \hline
\end{longtable}

\begin{longtable}{c}
    \caption{Perceptual attributes} \label{tab:perceptual} \\
    \hline \endfirsthead
        \hline
        Internal details \\ 
        \hline
        Verticality \\ 
        \hline
        Horizontalness \\ 
        \hline
        Left-diagonalness \\ 
        \hline
        Right-diagonalness \\ 
        \hline
        Aspect-ratio: skinny->fat \\ 
        \hline
        Prickiliness \\ 
        \hline
        Line curviness \\ 
        \hline
        3d curviness \\ 
        \hline
\end{longtable}

We also provide our split of attributes into \textit{spatial} (\cref{tab:spatial}) and \textit{non-spatial} (\cref{tab:non_spatial}) attributes, which is used in \cref{subsec:avg_rmse}. We define a spatial attribute to be anything relating to size, shape, color, material, subcomponents, being part of a larger entity, or anything else related to direct physicality. Non-spatial attributes were all remaining components. These typically involved higher level semantics.

\begin{center}
    \begin{longtable}{c}
    \caption{Spatial attributes}\label{tab:spatial} \\
    \hline \endfirsthead
    \hline \endhead

    \hline \textbf{List continued on the next page} \\ \hline
    \endfoot
    
    \hline
    \endlastfoot
        \hline
        Is it made of metal? \\ 
        \hline
        Is it made of plastic? \\ 
        \hline
        Is part of it made of glass? \\ 
        \hline
        Is it made of wood? \\ 
        \hline
        Is it shiny? \\ 
        \hline
        Can you see through it? \\ 
        \hline
        Is it colorful? \\ 
        \hline
        Is one more than one colored? \\ 
        \hline
        Is it always the same color(s)? \\ 
        \hline
        Is it white? \\ 
        \hline
        Is it red? \\ 
        \hline
        Is it orange? \\ 
        \hline
        Is it flesh-colored? \\ 
        \hline
        Is it yellow? \\ 
        \hline
        Is it green? \\ 
        \hline
        Is it blue? \\ 
        \hline
        Is it silver? \\ 
        \hline
        Is it brown? \\ 
        \hline
        Is it black? \\ 
        \hline
        Is it curved? \\ 
        \hline
        Is it straight? \\ 
        \hline
        Is it flat? \\ 
        \hline
        Does it have a front and a back? \\ 
        \hline
        Does it have a flat / straight top? \\ 
        \hline
        Does it have flat / straight sides? \\ 
        \hline
        Is taller than it is wide/long? \\ 
        \hline
        Is it long? \\ 
        \hline
        Is it pointed / sharp? \\ 
        \hline
        Is it tapered? \\ 
        \hline
        Is it round? \\ 
        \hline
        Does it have corners? \\ 
        \hline
        Is it symmetrical? \\ 
        \hline
        Is it hairy? \\ 
        \hline
        Is it fuzzy? \\ 
        \hline
        Is it clear? \\ 
        \hline
        Is it smooth? \\ 
        \hline
        Is it soft? \\ 
        \hline
        Is it heavy? \\ 
        \hline
        Is it lightweight? \\ 
        \hline
        Is it dense? \\ 
        \hline
        Is it slippery? \\ 
        \hline
        Can it bend? \\ 
        \hline
        Can it stretch? \\ 
        \hline
        Can it break? \\ 
        \hline
        Is it fragile? \\ 
        \hline
        Does it have parts? \\ 
        \hline
        Does it have moving parts? \\ 
        \hline
        Does it come in pairs? \\ 
        \hline
        Does it come in a bunch/pack? \\ 
        \hline
        Does it live in groups? \\ 
        \hline
        Is it part of something larger? \\ 
        \hline
        Does it contain something else? \\ 
        \hline
        Does it have internal structure? \\ 
        \hline
        Does it open? \\ 
        \hline
        Is it hollow? \\ 
        \hline
        Does it have a hard outer shell? \\ 
        \hline
        Does it have at least one hole? \\ 
        \hline
        Is it manufactured? \\ 
        \hline
        Does it come in different sizes? \\ 
        \hline
        Is it smaller than a golfball? \\ 
        \hline
        Is it bigger than a loaf of bread? \\ 
        \hline
        Is it bigger than a microwave oven? \\ 
        \hline
        Is it bigger than a bed? \\ 
        \hline
        Is it bigger than a car? \\ 
        \hline
        Is it bigger than a house? \\ 
        \hline
        Is it taller than a person? \\ 
        \hline
        Does it have a tail? \\ 
        \hline
        Does it have legs? \\ 
        \hline
        Does it have four legs? \\ 
        \hline
        Does it have feet? \\ 
        \hline
        Does it have paws? \\ 
        \hline
        Does it have claws? \\ 
        \hline
        Does it have horns / thorns / spikes? \\ 
        \hline
        Does it have hooves? \\ 
        \hline
        Does it have a face? \\ 
        \hline
        Does it have a backbone? \\ 
        \hline
        Does it have wings? \\ 
        \hline
        Does it have ears? \\ 
        \hline
        Does it have roots? \\ 
        \hline
        Does it have seeds? \\ 
        \hline
        Does it have leaves? \\ 
        \hline
        Does it have feathers? \\ 
        \hline
        Does it have some sort of nose? \\ 
        \hline
        Does it have a hard nose/beak? \\ 
        \hline
        Does it contain liquid? \\ 
        \hline
        Does it have wires or a cord? \\ 
        \hline
        Does it have writing on it? \\ 
        \hline
        Does it have wheels? \\ 
        \hline
        Does it roll? \\ 
        \hline
        Does it stand on two legs? \\ 
        \hline
        Is it mechanical? \\ 
        \hline
        Is it electronic? \\ 
        \hline
        Does it cast a shadow? \\ 
        \hline
        Can you hold it? \\ 
        \hline
        Can you hold it in one hand? \\ 
        \hline
        Can you pick it up? \\ 
        \hline
        Can you sit on it? \\ 
        \hline
        Can you ride on/in it? \\ 
        \hline
        Could you fit inside it? \\ 
        \hline
        Would you find it on a farm? \\ 
        \hline
        Would you find it in a school? \\ 
        \hline
        Would you find it in a zoo? \\ 
        \hline
        Would you find it in an office? \\ 
        \hline
        Would you find it in a restaurant? \\ 
        \hline
        Would you find in the bathroom? \\ 
        \hline
        Would you find it in a house? \\ 
        \hline
        Would you find it near a road? \\ 
        \hline
        Would you find it in a dump/landfill? \\ 
        \hline
        Would you find it in the forest? \\ 
        \hline
        Would you find it in a garden? \\ 
        \hline
        Would you find it in the sky? \\ 
        \hline
        Do you find it in space? \\ 
        \hline
        Does it live above ground? \\ 
        \hline
        Does it live in water? \\ 
        \hline
        Internal details \\ 
        \hline
        Verticality \\ 
        \hline
        Horizontalness \\ 
        \hline
        Left-diagonalness \\ 
        \hline
        Right-diagonalness \\ 
        \hline
        Aspect-ratio: skinny->fat \\ 
        \hline
        Prickiliness \\ 
        \hline
        Line curviness \\ 
        \hline
        3d curviness \\ 
        \hline

    \end{longtable}
\end{center}

\begin{center}
\begin{longtable}{c}
    \caption{Non-Spatial attributes}\label{tab:non_spatial} \\
    \hline \endfirsthead
    \hline \endhead
    
    \hline \textbf{List continued on the next page} \\ \hline
    \endfoot
    
    \hline
    \endlastfoot
        \hline
        Is it an animal? \\ 
        \hline
        Is it a body part? \\ 
        \hline
        Is it a building? \\ 
        \hline
        Is it a building part? \\ 
        \hline
        Is it clothing? \\ 
        \hline
        Is it furniture? \\ 
        \hline
        Is it an insect? \\ 
        \hline
        Is it a kitchen item? \\ 
        \hline
        Is it manmade? \\ 
        \hline
        Is it a tool? \\ 
        \hline
        Can you eat it? \\ 
        \hline
        Is it a vehicle? \\ 
        \hline
        Is it a person? \\ 
        \hline
        Is it a vegetable / plant? \\ 
        \hline
        Is it a fruit? \\ 
        \hline
        Does it change color? \\ 
        \hline
        Can it change shape? \\ 
        \hline
        Does it have a hard inside? \\ 
        \hline
        Is it alive? \\ 
        \hline
        Was it ever alive? \\ 
        \hline
        Is it a specific gender? \\ 
        \hline
        Was it invented? \\ 
        \hline
        Was it around 100 years ago? \\ 
        \hline
        Are there many varieties of it? \\ 
        \hline
        Does it grow? \\ 
        \hline
        Does it come from a plant? \\ 
        \hline
        Does it make a sound? \\ 
        \hline
        Does it make a nice sound? \\ 
        \hline
        Does it make sound continuously when active? \\ 
        \hline
        Is its job to make sounds? \\ 
        \hline
        Can it run? \\ 
        \hline
        Is it fast? \\ 
        \hline
        Can it fly? \\ 
        \hline
        Can it jump? \\ 
        \hline
        Can it float? \\ 
        \hline
        Can it swim? \\ 
        \hline
        Can it dig? \\ 
        \hline
        Can it climb trees? \\ 
        \hline
        Can it cause you pain? \\ 
        \hline
        Can it bite or sting? \\ 
        \hline
        Is it wild? \\ 
        \hline
        Is it a herbivore? \\ 
        \hline
        Is it a predator? \\ 
        \hline
        Is it warm blooded? \\ 
        \hline
        Is it a mammal? \\ 
        \hline
        Is it nocturnal? \\ 
        \hline
        Does it lay eggs? \\ 
        \hline
        Is it conscious? \\ 
        \hline
        Does it have feelings? \\ 
        \hline
        Is it smart? \\ 
        \hline
        Does it use electricity? \\ 
        \hline
        Can it keep you dry? \\ 
        \hline
        Does it provide protection? \\ 
        \hline
        Does it provide shade? \\ 
        \hline
        Do you see it daily? \\ 
        \hline
        Is it helpful? \\ 
        \hline
        Do you interact with it? \\ 
        \hline
        Can you touch it? \\ 
        \hline
        Would you avoid touching it? \\ 
        \hline
        Do you hold it to use it? \\ 
        \hline
        Can you play it? \\ 
        \hline
        Can you play with it? \\ 
        \hline
        Can you pet it? \\ 
        \hline
        Can you use it? \\ 
        \hline
        Do you use it daily? \\ 
        \hline
        Can you use it up? \\ 
        \hline
        Do you use it when cooking? \\ 
        \hline
        Is it used to carry things? \\ 
        \hline
        Can you control it? \\ 
        \hline
        Is it used for transportation? \\ 
        \hline
        Is it used in sports? \\ 
        \hline
        Do you wear it? \\ 
        \hline
        Can it be washed? \\ 
        \hline
        Is it cold? \\ 
        \hline
        Is it cool? \\ 
        \hline
        Is it warm? \\ 
        \hline
        Is it hot? \\ 
        \hline
        Is it unhealthy? \\ 
        \hline
        Is it hard to catch? \\ 
        \hline
        Can you peel it? \\ 
        \hline
        Can you walk on it? \\ 
        \hline
        Can you switch it on and off? \\ 
        \hline
        Can it be easily moved? \\ 
        \hline
        Do you drink from it? \\ 
        \hline
        Does it go in your mouth? \\ 
        \hline
        Is it tasty? \\ 
        \hline
        Is it used during meals? \\ 
        \hline
        Does it have a strong smell? \\ 
        \hline
        Does it smell good? \\ 
        \hline
        Does it smell bad? \\ 
        \hline
        Is it usually inside? \\ 
        \hline
        Is it usually outside? \\ 
        \hline
        Does it get wet? \\ 
        \hline
        Can it live out of water? \\ 
        \hline
        Do you take care of it? \\ 
        \hline
        Does it make you happy? \\ 
        \hline
        Do you love it? \\ 
        \hline
        Would you miss it if it were gone? \\ 
        \hline
        Is it scary? \\ 
        \hline
        Is it dangerous? \\ 
        \hline
        Is it friendly? \\ 
        \hline
        Is it rare? \\ 
        \hline
        Can you buy it? \\ 
        \hline
        Is it valuable? \\ 
        \hline
\end{longtable}
\end{center}

\bibliographystyle{splncs04}
\bibliography{main}

@misc{rombach2022highresolution,
      title={High-Resolution Image Synthesis with Latent Diffusion Models}, 
      author={Robin Rombach and Andreas Blattmann and Dominik Lorenz and Patrick Esser and Björn Ommer},
      year={2022},
      eprint={2112.10752},
      archivePrefix={arXiv},
      primaryClass={cs.CV}
}

@misc{po2023state,
      title={State of the Art on Diffusion Models for Visual Computing}, 
      author={Ryan Po and Wang Yifan and Vladislav Golyanik and Kfir Aberman and Jonathan T. Barron and Amit H. Bermano and Eric Ryan Chan and Tali Dekel and Aleksander Holynski and Angjoo Kanazawa and C. Karen Liu and Lingjie Liu and Ben Mildenhall and Matthias Nießner and Björn Ommer and Christian Theobalt and Peter Wonka and Gordon Wetzstein},
      year={2023},
      eprint={2310.07204},
      archivePrefix={arXiv},
      primaryClass={cs.AI}
}

@misc{luo2022understanding,
      title={Understanding Diffusion Models: A Unified Perspective}, 
      author={Calvin Luo},
      year={2022},
      eprint={2208.11970},
      archivePrefix={arXiv},
      primaryClass={cs.LG}
}

@misc{ronneberger2015unet,
      title={U-Net: Convolutional Networks for Biomedical Image Segmentation}, 
      author={Olaf Ronneberger and Philipp Fischer and Thomas Brox},
      year={2015},
      eprint={1505.04597},
      archivePrefix={arXiv},
      primaryClass={cs.CV}
}

@misc{sohldickstein2015deep,
      title={Deep Unsupervised Learning using Nonequilibrium Thermodynamics}, 
      author={Jascha Sohl-Dickstein and Eric A. Weiss and Niru Maheswaranathan and Surya Ganguli},
      year={2015},
      eprint={1503.03585},
      archivePrefix={arXiv},
      primaryClass={cs.LG}
}

@misc{song2022denoising,
      title={Denoising Diffusion Implicit Models}, 
      author={Jiaming Song and Chenlin Meng and Stefano Ermon},
      year={2022},
      eprint={2010.02502},
      archivePrefix={arXiv},
      primaryClass={cs.LG}
}

@misc{zhang2023adding,
      title={Adding Conditional Control to Text-to-Image Diffusion Models}, 
      author={Lvmin Zhang and Anyi Rao and Maneesh Agrawala},
      year={2023},
      eprint={2302.05543},
      archivePrefix={arXiv},
      primaryClass={cs.CV},
      url={https://arxiv.org/abs/2302.05543}
}

@misc{brooks2023instructpix2pix,
      title={InstructPix2Pix: Learning to Follow Image Editing Instructions}, 
      author={Tim Brooks and Aleksander Holynski and Alexei A. Efros},
      year={2023},
      eprint={2211.09800},
      archivePrefix={arXiv},
      primaryClass={cs.CV}
}

@misc{hong2023improving,
      title={Improving Sample Quality of Diffusion Models Using Self-Attention Guidance}, 
      author={Susung Hong and Gyuseong Lee and Wooseok Jang and Seungryong Kim},
      year={2023},
      eprint={2210.00939},
      archivePrefix={arXiv},
      primaryClass={cs.CV}
}

@misc{tang2022daam,
      title={What the DAAM: Interpreting Stable Diffusion Using Cross Attention}, 
      author={Raphael Tang and Linqing Liu and Akshat Pandey and Zhiying Jiang and Gefei Yang and Karun Kumar and Pontus Stenetorp and Jimmy Lin and Ferhan Ture},
      year={2022},
      eprint={2210.04885},
      archivePrefix={arXiv},
      primaryClass={cs.CV},
      url={https://arxiv.org/abs/2210.04885}
}

@misc{park2023understanding,
      title={Understanding the Latent Space of Diffusion Models through the Lens of Riemannian Geometry}, 
      author={Yong-Hyun Park and Mingi Kwon and Jaewoong Choi and Junghyo Jo and Youngjung Uh},
      year={2023},
      eprint={2307.12868},
      archivePrefix={arXiv},
      primaryClass={cs.CV}
}

@misc{hertz2022prompttoprompt,
      title={Prompt-to-Prompt Image Editing with Cross Attention Control}, 
      author={Amir Hertz and Ron Mokady and Jay Tenenbaum and Kfir Aberman and Yael Pritch and Daniel Cohen-Or},
      year={2022},
      eprint={2208.01626},
      archivePrefix={arXiv},
      primaryClass={cs.CV}
}

@misc{kim2023dense,
      title={Dense Text-to-Image Generation with Attention Modulation}, 
      author={Yunji Kim and Jiyoung Lee and Jin-Hwa Kim and Jung-Woo Ha and Jun-Yan Zhu},
      year={2023},
      eprint={2308.12964},
      archivePrefix={arXiv},
      primaryClass={cs.CV}
}

@misc{radford2021learning,
      title={Learning Transferable Visual Models From Natural Language Supervision}, 
      author={Alec Radford and Jong Wook Kim and Chris Hallacy and Aditya Ramesh and Gabriel Goh and Sandhini Agarwal and Girish Sastry and Amanda Askell and Pamela Mishkin and Jack Clark and Gretchen Krueger and Ilya Sutskever},
      year={2021},
      eprint={2103.00020},
      archivePrefix={arXiv},
      primaryClass={cs.CV}
}

@misc{lewis2023does,
      title={Does CLIP Bind Concepts? Probing Compositionality in Large Image Models}, 
      author={Martha Lewis and Nihal V. Nayak and Peilin Yu and Qinan Yu and Jack Merullo and Stephen H. Bach and Ellie Pavlick},
      year={2023},
      eprint={2212.10537},
      archivePrefix={arXiv},
      primaryClass={cs.CV}
}

@misc{thrush2022winoground,
      title={Winoground: Probing Vision and Language Models for Visio-Linguistic Compositionality}, 
      author={Tristan Thrush and Ryan Jiang and Max Bartolo and Amanpreet Singh and Adina Williams and Douwe Kiela and Candace Ross},
      year={2022},
      eprint={2204.03162},
      archivePrefix={arXiv},
      primaryClass={cs.CV}
}

@misc{schiappa2023probing,
      title={Probing Conceptual Understanding of Large Visual-Language Models}, 
      author={Madeline Chantry Schiappa and Michael Cogswell and Ajay Divakaran and Yogesh Singh Rawat},
      year={2023},
      eprint={2304.03659},
      archivePrefix={arXiv},
      primaryClass={cs.CV}
}

@misc{muttenthaler2023human,
      title={Human alignment of neural network representations}, 
      author={Lukas Muttenthaler and Jonas Dippel and Lorenz Linhardt and Robert A. Vandermeulen and Simon Kornblith},
      year={2023},
      eprint={2211.01201},
      archivePrefix={arXiv},
      primaryClass={cs.CV}
}

@misc{schuhmann2021laion400m,
      title={LAION-400M: Open Dataset of CLIP-Filtered 400 Million Image-Text Pairs}, 
      author={Christoph Schuhmann and Richard Vencu and Romain Beaumont and Robert Kaczmarczyk and Clayton Mullis and Aarush Katta and Theo Coombes and Jenia Jitsev and Aran Komatsuzaki},
      year={2021},
      eprint={2111.02114},
      archivePrefix={arXiv},
      primaryClass={cs.CV}
}

@article{Sudre2012,
  doi = {10.1016/j.neuroimage.2012.04.048},
  url = {https://doi.org/10.1016/j.neuroimage.2012.04.048},
  year = {2012},
  month = aug,
  publisher = {Elsevier {BV}},
  volume = {62},
  number = {1},
  pages = {451--463},
  author = {Gustavo Sudre and Dean Pomerleau and Mark Palatucci and Leila Wehbe and Alona Fyshe and Riitta Salmelin and Tom Mitchell},
  title = {Tracking neural coding of perceptual and semantic features of concrete nouns},
  journal = {{NeuroImage}}
}

@article{Jolliffe2016,
  title = {Principal component analysis: a review and recent developments},
  volume = {374},
  ISSN = {1471-2962},
  url = {http://dx.doi.org/10.1098/rsta.2015.0202},
  DOI = {10.1098/rsta.2015.0202},
  number = {2065},
  journal = {Philosophical Transactions of the Royal Society A: Mathematical,  Physical and Engineering Sciences},
  publisher = {The Royal Society},
  author = {Jolliffe,  Ian T. and Cadima,  Jorge},
  year = {2016},
  month = apr,
  pages = {20150202}
}

@misc{han2012mining,
  address = {Waltham, Mass.},
  author = {Han, Jiawei and Kamber, Micheline and Pei, Jian},
  description = {Data Mining: Concepts and Techniques (Morgan Kaufmann Series in Data Management Systems): Amazon.de: Jiawei Han, Micheline Kamber, Jian Pei: Englische Bücher},
  isbn = {0123814790},
  publisher = {Morgan Kaufmann Publishers},
  refid = {818321921},
  title = {Data mining concepts and techniques, third edition},
  url = {http://www.amazon.de/Data-Mining-Concepts-Techniques-Management/dp/0123814790/ref=tmm_hrd_title_0?ie=UTF8&qid=1366039033&sr=1-1},
  year = 2012
}

@misc{kang2023scaling,
      title={Scaling up GANs for Text-to-Image Synthesis}, 
      author={Minguk Kang and Jun-Yan Zhu and Richard Zhang and Jaesik Park and Eli Shechtman and Sylvain Paris and Taesung Park},
      year={2023},
      eprint={2303.05511},
      archivePrefix={arXiv},
      primaryClass={cs.CV}
}

@misc{kwon2023diffusion,
      title={Diffusion Models already have a Semantic Latent Space}, 
      author={Mingi Kwon and Jaeseok Jeong and Youngjung Uh},
      year={2023},
      eprint={2210.10960},
      archivePrefix={arXiv},
      primaryClass={cs.CV}
}

@article{Hentschel2022,
  doi = {10.3389/frai.2022.976235},
  url = {https://doi.org/10.3389/frai.2022.976235},
  year = {2022},
  month = nov,
  publisher = {Frontiers Media {SA}},
  volume = {5},
  author = {Simon Hentschel and Konstantin Kobs and Andreas Hotho},
  title = {{CLIP} knows image aesthetics},
  journal = {Frontiers in Artificial Intelligence}
}

@misc{ramesh2021zeroshot,
      title={Zero-Shot Text-to-Image Generation}, 
      author={Aditya Ramesh and Mikhail Pavlov and Gabriel Goh and Scott Gray and Chelsea Voss and Alec Radford and Mark Chen and Ilya Sutskever},
      year={2021},
      eprint={2102.12092},
      archivePrefix={arXiv},
      primaryClass={cs.CV}
}

@misc{ramesh2022hierarchical,
      title={Hierarchical Text-Conditional Image Generation with CLIP Latents}, 
      author={Aditya Ramesh and Prafulla Dhariwal and Alex Nichol and Casey Chu and Mark Chen},
      year={2022},
      eprint={2204.06125},
      archivePrefix={arXiv},
      primaryClass={cs.CV}
}

@misc{belinkov2021probing,
      title={Probing Classifiers: Promises, Shortcomings, and Advances}, 
      author={Yonatan Belinkov},
      year={2021},
      eprint={2102.12452},
      archivePrefix={arXiv},
      primaryClass={cs.CL}
}

@book{hastie01statisticallearning,
  address = {New York, NY, USA},
  author = {Hastie, Trevor and Tibshirani, Robert and Friedman, Jerome},
  publisher = {Springer New York Inc.},
  series = {Springer Series in Statistics},
  title = {The Elements of Statistical Learning},
  year = 2001
}

@INPROCEEDINGS{5360332,
  author={Ojala, Markus and Garriga, Gemma C.},
  booktitle={2009 Ninth IEEE International Conference on Data Mining}, 
  title={Permutation Tests for Studying Classifier Performance}, 
  year={2009},
  volume={},
  number={},
  pages={908-913},
  doi={10.1109/ICDM.2009.108}}

@book{Good2000,
  title = {Permutation Tests},
  ISBN = {9781475732351},
  ISSN = {0172-7397},
  url = {http://dx.doi.org/10.1007/978-1-4757-3235-1},
  DOI = {10.1007/978-1-4757-3235-1},
  journal = {Springer Series in Statistics},
  publisher = {Springer New York},
  author = {Good,  Phillip},
  year = {2000}
}

@article{Jeffers1967,
  title = {Two Case Studies in the Application of Principal Component Analysis},
  volume = {16},
  ISSN = {0035-9254},
  url = {http://dx.doi.org/10.2307/2985919},
  DOI = {10.2307/2985919},
  number = {3},
  journal = {Applied Statistics},
  publisher = {JSTOR},
  author = {Jeffers,  J. N. R.},
  year = {1967},
  pages = {225}
}

@misc{alain2018understanding,
      title={Understanding intermediate layers using linear classifier probes}, 
      author={Guillaume Alain and Yoshua Bengio},
      year={2018},
      eprint={1610.01644},
      archivePrefix={arXiv},
      primaryClass={stat.ML}
}

@inproceedings{kohn2015,
    author = {Köhn, Arne},
    year = {2015},
    month = {01},
    pages = {2067-2073},
    title = {What’s in an Embedding? Analyzing Word Embeddings through Multilingual Evaluation},
    booktitle = {Proceedings of the 2015 Conference on Empirical Methods in Natural Language Processing},
    doi = {10.18653/v1/D15-1246}
}

@inproceedings{gupta2015,
    author = {Gupta, Abhijeet and Boleda, Gemma and Baroni, Marco and Padó, Sebastian},
    year = {2015},
    month = {09},
    pages = {},
    title = {Distributional vectors encode referential attributes},
    booktitle = {Proceedings of the 2015 Conference on Empirical Methods in Natural Language Processing},
    doi = {10.18653/v1/D15-1002}
}

@inproceedings{shi2016,
    author = {Shi, Xing and Padhi, Inkit and Knight, Kevin},
    year = {2016},
    month = {01},
    pages = {1526-1534},
    title = {Does String-Based Neural MT Learn Source Syntax?},
    booktitle={Conference on Empirical Methods in Natural Language Processing},
    doi = {10.18653/v1/D16-1159}
}

@misc{peebles2023scalable,
      title={Scalable Diffusion Models with Transformers}, 
      author={William Peebles and Saining Xie},
      year={2023},
      eprint={2212.09748},
      archivePrefix={arXiv},
      primaryClass={cs.CV}
}

@misc{zhang2023texttoimage,
      title={Text-to-image Diffusion Models in Generative AI: A Survey}, 
      author={Chenshuang Zhang and Chaoning Zhang and Mengchun Zhang and In So Kweon},
      year={2023},
      eprint={2303.07909},
      archivePrefix={arXiv},
      primaryClass={cs.CV}
}

@Inbook{Refaeilzadeh2009,
    author="Refaeilzadeh, Payam and Tang, Lei and Liu, Huan",
    title="Cross-Validation",
    bookTitle="Encyclopedia of Database Systems",
    year="2009",
    publisher="Springer US",
    address="Boston, MA",
    pages="532--538",
    isbn="978-0-387-39940-9",
    doi="10.1007/978-0-387-39940-9_565",
    url="https://doi.org/10.1007/978-0-387-39940-9_565"
}

@article{wang2022fmri,
  title={An fmri dataset for concept representation with semantic feature annotations},
  author={Wang, Shaonan and Zhang, Yunhao and Zhang, Xiaohan and Sun, Jingyuan and Lin, Nan and Zhang, Jiajun and Zong, Chengqing},
  journal={Scientific Data},
  volume={9},
  number={1},
  pages={721},
  year={2022},
  publisher={Nature Publishing Group UK London}
}

@article{mcrae2005semantic,
  title={Semantic feature production norms for a large set of living and nonliving things},
  author={McRae, Ken and Cree, George S and Seidenberg, Mark S and McNorgan, Chris},
  journal={Behavior research methods},
  volume={37},
  number={4},
  pages={547--559},
  year={2005},
  publisher={Springer}
}

@misc{ray2023cola,
      title={COLA: A Benchmark for Compositional Text-to-image Retrieval}, 
      author={Arijit Ray and Filip Radenovic and Abhimanyu Dubey and Bryan A. Plummer and Ranjay Krishna and Kate Saenko},
      year={2023},
      eprint={2305.03689},
      archivePrefix={arXiv},
      primaryClass={cs.CV}
}

@misc{yun2023visionlanguage,
      title={Do Vision-Language Pretrained Models Learn Composable Primitive Concepts?}, 
      author={Tian Yun and Usha Bhalla and Ellie Pavlick and Chen Sun},
      year={2023},
      eprint={2203.17271},
      archivePrefix={arXiv},
      primaryClass={cs.CV}
}

@article{Stone1974,
  title = {Cross‐Validatory Choice and Assessment of Statistical Predictions},
  volume = {36},
  ISSN = {2517-6161},
  url = {http://dx.doi.org/10.1111/j.2517-6161.1974.tb00994.x},
  DOI = {10.1111/j.2517-6161.1974.tb00994.x},
  number = {2},
  journal = {Journal of the Royal Statistical Society: Series B (Methodological)},
  publisher = {Wiley},
  author = {Stone,  M.},
  year = {1974},
  month = jan,
  pages = {111–133}
}

@article{cawley2010,
author = {Cawley, Gavin and Talbot, Nicola},
year = {2010},
month = {07},
pages = {2079-2107},
title = {On Over-fitting in Model Selection and Subsequent Selection Bias in Performance Evaluation},
volume = {11},
journal = {Journal of Machine Learning Research}
}

@article{Xu2017-rq,
  title    = "The differences and similarities between two-sample T-test and
              paired T-test",
  author   = "Xu, Manfei and Fralick, Drew and Zheng, Julia Z and Wang, Bokai
              and Tu, Xin M and Feng, Changyong",
  journal  = "Shanghai Arch. Psychiatry",
  volume   =  29,
  number   =  3,
  pages    = "184--188",
  month    =  jun,
  year     =  2017,
  language = "en"
}

@misc{balaji2023ediffi,
      title={eDiff-I: Text-to-Image Diffusion Models with an Ensemble of Expert Denoisers}, 
      author={Yogesh Balaji and Seungjun Nah and Xun Huang and Arash Vahdat and Jiaming Song and Qinsheng Zhang and Karsten Kreis and Miika Aittala and Timo Aila and Samuli Laine and Bryan Catanzaro and Tero Karras and Ming-Yu Liu},
      year={2023},
      eprint={2211.01324},
      archivePrefix={arXiv},
      primaryClass={cs.CV}
}

@misc{saharia2022photorealistic,
      title={Photorealistic Text-to-Image Diffusion Models with Deep Language Understanding}, 
      author={Chitwan Saharia and William Chan and Saurabh Saxena and Lala Li and Jay Whang and Emily Denton and Seyed Kamyar Seyed Ghasemipour and Burcu Karagol Ayan and S. Sara Mahdavi and Rapha Gontijo Lopes and Tim Salimans and Jonathan Ho and David J Fleet and Mohammad Norouzi},
      year={2022},
      eprint={2205.11487},
      archivePrefix={arXiv},
      primaryClass={cs.CV}
}

@misc{nichol2022glide,
      title={GLIDE: Towards Photorealistic Image Generation and Editing with Text-Guided Diffusion Models}, 
      author={Alex Nichol and Prafulla Dhariwal and Aditya Ramesh and Pranav Shyam and Pamela Mishkin and Bob McGrew and Ilya Sutskever and Mark Chen},
      year={2022},
      eprint={2112.10741},
      archivePrefix={arXiv},
      primaryClass={cs.CV}
}

@misc{li2023gligen,
      title={GLIGEN: Open-Set Grounded Text-to-Image Generation}, 
      author={Yuheng Li and Haotian Liu and Qingyang Wu and Fangzhou Mu and Jianwei Yang and Jianfeng Gao and Chunyuan Li and Yong Jae Lee},
      year={2023},
      eprint={2301.07093},
      archivePrefix={arXiv},
      primaryClass={cs.CV}
}

@misc{arkhipkin2023kandinsky,
      title={Kandinsky 3.0 Technical Report}, 
      author={Vladimir Arkhipkin and Andrei Filatov and Viacheslav Vasilev and Anastasia Maltseva and Said Azizov and Igor Pavlov and Julia Agafonova and Andrey Kuznetsov and Denis Dimitrov},
      year={2023},
      eprint={2312.03511},
      archivePrefix={arXiv},
      primaryClass={cs.CV}
}

@Inbook{Haddad2009,
    author="Haddad, Caroline N.",
    title="Cholesky Factorization",
    bookTitle="Encyclopedia of Optimization",
    year="2009",
    publisher="Springer US",
    address="Boston, MA",
    pages="374--377",
    isbn="978-0-387-74759-0",
    doi="10.1007/978-0-387-74759-0_67",
    url="https://doi.org/10.1007/978-0-387-74759-0_67"
}

@misc{bie2023renaissance,
      title={RenAIssance: A Survey into AI Text-to-Image Generation in the Era of Large Model}, 
      author={Fengxiang Bie and Yibo Yang and Zhongzhu Zhou and Adam Ghanem and Minjia Zhang and Zhewei Yao and Xiaoxia Wu and Connor Holmes and Pareesa Golnari and David A. Clifton and Yuxiong He and Dacheng Tao and Shuaiwen Leon Song},
      year={2023},
      eprint={2309.00810},
      archivePrefix={arXiv},
      primaryClass={cs.CV}
}

@misc{aw2023training,
      title={Training language models to summarize narratives improves brain alignment}, 
      author={Khai Loong Aw and Mariya Toneva},
      year={2023},
      eprint={2212.10898},
      archivePrefix={arXiv},
      primaryClass={cs.CL}
}

@misc{abdou2021does,
      title={Does injecting linguistic structure into language models lead to better alignment with brain recordings?}, 
      author={Mostafa Abdou and Ana Valeria Gonzalez and Mariya Toneva and Daniel Hershcovich and Anders Søgaard},
      year={2021},
      eprint={2101.12608},
      archivePrefix={arXiv},
      primaryClass={cs.CL}
}

@article{Moss2023,
  title = {Is it ethical to use Mechanical Turk for behavioral research? Relevant data from a representative survey of MTurk participants and wages},
  volume = {55},
  ISSN = {1554-3528},
  url = {http://dx.doi.org/10.3758/s13428-022-02005-0},
  DOI = {10.3758/s13428-022-02005-0},
  number = {8},
  journal = {Behavior Research Methods},
  publisher = {Springer Science and Business Media LLC},
  author = {Moss,  Aaron J. and Rosenzweig,  Cheskie and Robinson,  Jonathan and Jaffe,  Shalom N. and Litman,  Leib},
  year = {2023},
  month = may,
  pages = {4048–4067}
}

\clearpage

\end{document}